\documentclass[conference]{IEEEtran}
\IEEEoverridecommandlockouts

\usepackage{cite}
\usepackage{amsmath,amssymb,amsfonts}
\usepackage{graphicx}
\usepackage{booktabs}
\usepackage{multirow}
\usepackage{array}
\usepackage[table]{xcolor}
\usepackage{tikz}
\usetikzlibrary{arrows.meta,positioning,calc,fit,shapes.geometric}
\usepackage{fontawesome5} 
\usepackage[hidelinks]{hyperref}
\usepackage{url}

\newif\ifdraft
\drafttrue
\ifdraft
  \newcommand{\res}[1]{\textcolor{blue}{[#1]}}
  \newcommand{\todo}[1]{\textcolor{red}{\textbf{[TODO:} #1\textbf{]}}}
\else
  \newcommand{\res}[1]{}
  \newcommand{\todo}[1]{}
\fi

\newcommand{\resdir}{results/}
\makeatletter
\newcommand{\setres}[2]{\expandafter\gdef\csname res@#1\endcsname{#2}}
\newcommand{\rv}[1]{\ifcsname res@#1\endcsname\csname res@#1\endcsname\else\textcolor{red}{\textbf{??}}\fi}
\newcommand{\defrows}[2]{\IfFileExists{\resdir #1.tex}%
  {\expandafter\def\csname rows@#1\endcsname{\@@input \resdir #1.tex }}%
  {\expandafter\def\csname rows@#1\endcsname{\multicolumn{#2}{c}{\textcolor{red}{results pending}}\\}}}
\newcommand{\resrows}[1]{\csname rows@#1\endcsname}
\makeatother
\IfFileExists{\resdir results_auto.tex}{
\setres{F1.ext}{0}
\setres{F1.n}{60}
\setres{F1.silent}{0}
\setres{F1.tasks}{58}
\setres{F1.unchanged}{50}
\setres{F2.ext}{83}
\setres{F2.n}{18}
\setres{F2.silent}{0}
\setres{F2.tasks}{18}
\setres{F2.unchanged}{100}
\setres{F3.ext}{25}
\setres{F3.n}{60}
\setres{F3.silent}{0}
\setres{F3.tasks}{48}
\setres{F3.unchanged}{100}
\setres{F4.ext}{23}
\setres{F4.n}{60}
\setres{F4.silent}{0}
\setres{F4.tasks}{56}
\setres{F4.unchanged}{100}
\setres{F5.ext}{39}
\setres{F5.n}{38}
\setres{F5.silent}{100}
\setres{F5.tasks}{35}
\setres{F5.unchanged}{100}
\setres{F6.ext}{0}
\setres{F6.n}{60}
\setres{F6.silent}{0}
\setres{F6.tasks}{60}
\setres{F6.unchanged}{0}
\setres{F7.ext}{0}
\setres{F7.n}{60}
\setres{F7.silent}{0}
\setres{F7.tasks}{57}
\setres{F7.unchanged}{72}
\setres{F8.ext}{2}
\setres{F8.n}{60}
\setres{F8.silent}{0}
\setres{F8.tasks}{45}
\setres{F8.unchanged}{100}
\setres{H1.ci}{[$-$0.03, 0.00]}
\setres{H1.ci.alt}{[$-$0.03, 0.00]}
\setres{H1.est}{$-$0.01}
\setres{H1.family}{confirmatory}
\setres{H1.n}{1,854}
\setres{H1.p}{$<$0.001}
\setres{H1.peq}{$<$\,0.001}
\setres{H1.pholm}{$<$0.001}
\setres{H2.ci}{[0.14, 0.29]}
\setres{H2.ci.alt}{[0.15, 0.28]}
\setres{H2.est}{0.22}
\setres{H2.family}{confirmatory}
\setres{H2.n}{1,416}
\setres{H2.p}{1.000}
\setres{H2.peq}{$=$\,1.000}
\setres{H2.pholm}{1.000}
\setres{H3.ci}{[0.07, 0.24]}
\setres{H3.ci.alt}{[0.08, 0.23]}
\setres{H3.est}{0.16}
\setres{H3.family}{exploratory}
\setres{H3.n}{2,496}
\setres{H3.p}{0.993}
\setres{H3.peq}{$=$\,0.993}
\setres{H3.pholm}{0.993}
\setres{H4.ci}{[0.57, 0.71]}
\setres{H4.ci.alt}{[0.56, 0.71]}
\setres{H4.est}{0.64}
\setres{H4.family}{confirmatory}
\setres{H4.n}{720}
\setres{H4.p}{$<$0.001}
\setres{H4.peq}{$<$\,0.001}
\setres{H4.pholm}{$<$0.001}
\setres{H4.view}{V3n}
\setres{H6.ci}{[$-$0.14, $-$0.02]}
\setres{H6.ci.alt}{[$-$0.14, $-$0.02]}
\setres{H6.est}{$-$0.08}
\setres{H6.family}{confirmatory}
\setres{H6.n}{832}
\setres{H6.p}{$<$0.001}
\setres{H6.peq}{$<$\,0.001}
\setres{H6.pholm}{$<$0.001}
\setres{H7.cell.gemma.gpt-oss}{91}
\setres{H7.cell.gemma.gpt-oss.n}{358}
\setres{H7.cell.gemma.qwen}{50}
\setres{H7.cell.gemma.qwen.n}{120}
\setres{H7.cell.gpt-oss.gpt-oss}{88}
\setres{H7.cell.gpt-oss.gpt-oss.n}{358}
\setres{H7.cell.gpt-oss.qwen}{77}
\setres{H7.cell.gpt-oss.qwen.n}{120}
\setres{H7.cell.qwen.gpt-oss}{95}
\setres{H7.cell.qwen.gpt-oss.n}{358}
\setres{H7.cell.qwen.qwen}{63}
\setres{H7.cell.qwen.qwen.n}{120}
\setres{H7.ci}{[$-$0.15, $-$0.05]}
\setres{H7.ci.alt}{[$-$0.15, $-$0.05]}
\setres{H7.est}{$-$0.10}
\setres{H7.family}{exploratory}
\setres{H7.n}{956}
\setres{H7.p}{$<$0.001}
\setres{H7.peq}{$<$\,0.001}
\setres{H7.pholm}{$<$0.001}
\setres{H7.strat.ci}{[$-$0.07, 0.08]}
\setres{H7.strat.d.ci}{[$-$0.05, 0.12]}
\setres{H7.strat.d.est}{0.04}
\setres{H7.strat.d.peq}{$=$\,0.807}
\setres{H7.strat.est}{0.01}
\setres{H7.strat.peq}{$=$\,0.569}
\setres{S3.ext}{83}
\setres{S3.faults}{9 dose magnitude, 2 ignored observation and 1 wrong code}
\setres{S3.tasks}{11}
\setres{S3.tasktypes}{2}
\setres{S3.unchanged}{100}
\setres{agents.list}{gpt-oss 120B and Qwen3 32B}
\setres{agents.pass.list}{gpt-oss 120B 48\%, Qwen3 32B 45\%}
\setres{auroc.V3I2}{0.57}
\setres{auroc.flag.max}{0.91}
\setres{auroc.flag.min}{0.84}
\setres{auroc.gemma-3-27b.flag}{0.91}
\setres{auroc.gemma-3-27b.side}{0.94}
\setres{auroc.gemma-3-27b.stated}{0.52}
\setres{auroc.gpt-oss-120b.flag}{0.88}
\setres{auroc.gpt-oss-120b.side}{0.93}
\setres{auroc.gpt-oss-120b.stated}{0.63}
\setres{auroc.gpt-oss-20b.flag}{0.85}
\setres{auroc.gpt-oss-20b.side}{0.81}
\setres{auroc.gpt-oss-20b.stated}{0.62}
\setres{auroc.medgemma-27b.flag}{0.87}
\setres{auroc.medgemma-27b.side}{0.89}
\setres{auroc.medgemma-27b.stated}{0.55}
\setres{auroc.qwen3-32b.flag}{0.91}
\setres{auroc.qwen3-32b.side}{0.93}
\setres{auroc.qwen3-32b.stated}{0.66}
\setres{auroc.qwen3-8b.flag}{0.84}
\setres{auroc.qwen3-8b.side}{0.86}
\setres{auroc.qwen3-8b.stated}{0.46}
\setres{auroc.side.max}{0.94}
\setres{auroc.side.min}{0.81}
\setres{auroc.stated.max}{0.66}
\setres{auroc.stated.min}{0.46}
\setres{bootstrap.B}{10,000}
\setres{c0.V1I1.d}{$-$0.03}
\setres{c0.V1I1.d.ci}{[$-$0.07, 0.02]}
\setres{c0.V1I1.flag}{41}
\setres{c0.V1I1.pflag}{44}
\setres{c0.V1I2.d}{$-$0.02}
\setres{c0.V1I2.d.ci}{[$-$0.07, 0.03]}
\setres{c0.V1I2.flag}{41}
\setres{c0.V1I2.pflag}{43}
\setres{c0.V2I1.d}{$-$0.07}
\setres{c0.V2I1.d.ci}{[$-$0.12, $-$0.01]}
\setres{c0.V2I1.flag}{48}
\setres{c0.V2I1.pflag}{55}
\setres{c0.V2I2.d}{$-$0.02}
\setres{c0.V2I2.d.ci}{[$-$0.08, 0.03]}
\setres{c0.V2I2.flag}{44}
\setres{c0.V2I2.pflag}{47}
\setres{c0.V3I1.d}{$-$0.01}
\setres{c0.V3I1.d.ci}{[$-$0.02, 0.01]}
\setres{c0.V3I1.flag}{7}
\setres{c0.V3I1.pflag}{7}
\setres{c0.V3I2.d}{0.00}
\setres{c0.V3I2.d.ci}{[$-$0.02, 0.02]}
\setres{c0.V3I2.flag}{7}
\setres{c0.V3I2.pflag}{7}
\setres{c0.V3I3.d}{0.01}
\setres{c0.V3I3.d.ci}{[$-$0.02, 0.04]}
\setres{c0.V3I3.flag}{7}
\setres{c0.V3I3.pflag}{6}
\setres{c0.V3nI2.d}{0.01}
\setres{c0.V3nI2.d.ci}{[$-$0.01, 0.04]}
\setres{c0.V3nI2.flag}{8}
\setres{c0.V3nI2.pflag}{7}
\setres{cluster.alt}{parent trajectory}
\setres{cluster.alt.units}{parent trajectories}
\setres{cluster.unit}{task}
\setres{cluster.units}{tasks}
\setres{cm.V1I1}{11}
\setres{cm.V1I1.F1}{n/a}
\setres{cm.V1I1.F1.ci}{}
\setres{cm.V1I1.F1.n}{0}
\setres{cm.V1I1.F2}{11}
\setres{cm.V1I1.F2.ci}{[2, 21]}
\setres{cm.V1I1.F2.n}{9}
\setres{cm.V1I1.F3}{n/a}
\setres{cm.V1I1.F3.ci}{}
\setres{cm.V1I1.F3.n}{0}
\setres{cm.V1I1.F4}{17}
\setres{cm.V1I1.F4.ci}{[0, 33]}
\setres{cm.V1I1.F4.n}{2}
\setres{cm.V1I1.F5}{n/a}
\setres{cm.V1I1.F5.ci}{}
\setres{cm.V1I1.F5.n}{0}
\setres{cm.V1I1.F6}{n/a}
\setres{cm.V1I1.F6.ci}{}
\setres{cm.V1I1.F6.n}{0}
\setres{cm.V1I1.F7}{n/a}
\setres{cm.V1I1.F7.ci}{}
\setres{cm.V1I1.F7.n}{0}
\setres{cm.V1I1.F8}{0}
\setres{cm.V1I1.F8.ci}{[0, 0]}
\setres{cm.V1I1.F8.n}{1}
\setres{cm.V1I1.ci}{[3, 21]}
\setres{cm.V1I2}{21}
\setres{cm.V1I2.F1}{n/a}
\setres{cm.V1I2.F1.ci}{}
\setres{cm.V1I2.F1.n}{0}
\setres{cm.V1I2.F2}{20}
\setres{cm.V1I2.F2.ci}{[11, 29]}
\setres{cm.V1I2.F2.n}{9}
\setres{cm.V1I2.F3}{n/a}
\setres{cm.V1I2.F3.ci}{}
\setres{cm.V1I2.F3.n}{0}
\setres{cm.V1I2.F4}{33}
\setres{cm.V1I2.F4.ci}{[17, 50]}
\setres{cm.V1I2.F4.n}{2}
\setres{cm.V1I2.F5}{n/a}
\setres{cm.V1I2.F5.ci}{}
\setres{cm.V1I2.F5.n}{0}
\setres{cm.V1I2.F6}{n/a}
\setres{cm.V1I2.F6.ci}{}
\setres{cm.V1I2.F6.n}{0}
\setres{cm.V1I2.F7}{n/a}
\setres{cm.V1I2.F7.ci}{}
\setres{cm.V1I2.F7.n}{0}
\setres{cm.V1I2.F8}{0}
\setres{cm.V1I2.F8.ci}{[0, 0]}
\setres{cm.V1I2.F8.n}{1}
\setres{cm.V1I2.ci}{[12, 31]}
\setres{cm.V2I1}{0}
\setres{cm.V2I1.F1}{n/a}
\setres{cm.V2I1.F1.ci}{}
\setres{cm.V2I1.F1.n}{0}
\setres{cm.V2I1.F2}{0}
\setres{cm.V2I1.F2.ci}{[0, 0]}
\setres{cm.V2I1.F2.n}{9}
\setres{cm.V2I1.F3}{n/a}
\setres{cm.V2I1.F3.ci}{}
\setres{cm.V2I1.F3.n}{0}
\setres{cm.V2I1.F4}{0}
\setres{cm.V2I1.F4.ci}{[0, 0]}
\setres{cm.V2I1.F4.n}{2}
\setres{cm.V2I1.F5}{n/a}
\setres{cm.V2I1.F5.ci}{}
\setres{cm.V2I1.F5.n}{0}
\setres{cm.V2I1.F6}{n/a}
\setres{cm.V2I1.F6.ci}{}
\setres{cm.V2I1.F6.n}{0}
\setres{cm.V2I1.F7}{n/a}
\setres{cm.V2I1.F7.ci}{}
\setres{cm.V2I1.F7.n}{0}
\setres{cm.V2I1.F8}{0}
\setres{cm.V2I1.F8.ci}{[0, 0]}
\setres{cm.V2I1.F8.n}{1}
\setres{cm.V2I1.ci}{[0, 0]}
\setres{cm.V2I2}{0}
\setres{cm.V2I2.F1}{n/a}
\setres{cm.V2I2.F1.ci}{}
\setres{cm.V2I2.F1.n}{0}
\setres{cm.V2I2.F2}{0}
\setres{cm.V2I2.F2.ci}{[0, 0]}
\setres{cm.V2I2.F2.n}{9}
\setres{cm.V2I2.F3}{n/a}
\setres{cm.V2I2.F3.ci}{}
\setres{cm.V2I2.F3.n}{0}
\setres{cm.V2I2.F4}{0}
\setres{cm.V2I2.F4.ci}{[0, 0]}
\setres{cm.V2I2.F4.n}{2}
\setres{cm.V2I2.F5}{n/a}
\setres{cm.V2I2.F5.ci}{}
\setres{cm.V2I2.F5.n}{0}
\setres{cm.V2I2.F6}{n/a}
\setres{cm.V2I2.F6.ci}{}
\setres{cm.V2I2.F6.n}{0}
\setres{cm.V2I2.F7}{n/a}
\setres{cm.V2I2.F7.ci}{}
\setres{cm.V2I2.F7.n}{0}
\setres{cm.V2I2.F8}{0}
\setres{cm.V2I2.F8.ci}{[0, 0]}
\setres{cm.V2I2.F8.n}{1}
\setres{cm.V2I2.ci}{[0, 0]}
\setres{cm.V3I1}{8}
\setres{cm.V3I1.F1}{n/a}
\setres{cm.V3I1.F1.ci}{}
\setres{cm.V3I1.F1.n}{0}
\setres{cm.V3I1.F2}{11}
\setres{cm.V3I1.F2.ci}{[6, 17]}
\setres{cm.V3I1.F2.n}{9}
\setres{cm.V3I1.F3}{n/a}
\setres{cm.V3I1.F3.ci}{}
\setres{cm.V3I1.F3.n}{0}
\setres{cm.V3I1.F4}{0}
\setres{cm.V3I1.F4.ci}{[0, 0]}
\setres{cm.V3I1.F4.n}{2}
\setres{cm.V3I1.F5}{n/a}
\setres{cm.V3I1.F5.ci}{}
\setres{cm.V3I1.F5.n}{0}
\setres{cm.V3I1.F6}{n/a}
\setres{cm.V3I1.F6.ci}{}
\setres{cm.V3I1.F6.n}{0}
\setres{cm.V3I1.F7}{n/a}
\setres{cm.V3I1.F7.ci}{}
\setres{cm.V3I1.F7.n}{0}
\setres{cm.V3I1.F8}{0}
\setres{cm.V3I1.F8.ci}{[0, 0]}
\setres{cm.V3I1.F8.n}{1}
\setres{cm.V3I1.ci}{[4, 13]}
\setres{cm.V3I2}{6}
\setres{cm.V3I2.F1}{n/a}
\setres{cm.V3I2.F1.ci}{}
\setres{cm.V3I2.F1.n}{0}
\setres{cm.V3I2.F2}{7}
\setres{cm.V3I2.F2.ci}{[2, 13]}
\setres{cm.V3I2.F2.n}{9}
\setres{cm.V3I2.F3}{n/a}
\setres{cm.V3I2.F3.ci}{}
\setres{cm.V3I2.F3.n}{0}
\setres{cm.V3I2.F4}{0}
\setres{cm.V3I2.F4.ci}{[0, 0]}
\setres{cm.V3I2.F4.n}{2}
\setres{cm.V3I2.F5}{n/a}
\setres{cm.V3I2.F5.ci}{}
\setres{cm.V3I2.F5.n}{0}
\setres{cm.V3I2.F6}{n/a}
\setres{cm.V3I2.F6.ci}{}
\setres{cm.V3I2.F6.n}{0}
\setres{cm.V3I2.F7}{n/a}
\setres{cm.V3I2.F7.ci}{}
\setres{cm.V3I2.F7.n}{0}
\setres{cm.V3I2.F8}{0}
\setres{cm.V3I2.F8.ci}{[0, 0]}
\setres{cm.V3I2.F8.n}{1}
\setres{cm.V3I2.ci}{[1, 11]}
\setres{cm.V3I3}{3}
\setres{cm.V3I3.F1}{n/a}
\setres{cm.V3I3.F1.ci}{}
\setres{cm.V3I3.F1.n}{0}
\setres{cm.V3I3.F2}{2}
\setres{cm.V3I3.F2.ci}{[0, 6]}
\setres{cm.V3I3.F2.n}{9}
\setres{cm.V3I3.F3}{n/a}
\setres{cm.V3I3.F3.ci}{}
\setres{cm.V3I3.F3.n}{0}
\setres{cm.V3I3.F4}{8}
\setres{cm.V3I3.F4.ci}{[0, 17]}
\setres{cm.V3I3.F4.n}{2}
\setres{cm.V3I3.F5}{n/a}
\setres{cm.V3I3.F5.ci}{}
\setres{cm.V3I3.F5.n}{0}
\setres{cm.V3I3.F6}{n/a}
\setres{cm.V3I3.F6.ci}{}
\setres{cm.V3I3.F6.n}{0}
\setres{cm.V3I3.F7}{n/a}
\setres{cm.V3I3.F7.ci}{}
\setres{cm.V3I3.F7.n}{0}
\setres{cm.V3I3.F8}{0}
\setres{cm.V3I3.F8.ci}{[0, 0]}
\setres{cm.V3I3.F8.n}{1}
\setres{cm.V3I3.ci}{[0, 7]}
\setres{cm.V3nI2}{3}
\setres{cm.V3nI2.F1}{n/a}
\setres{cm.V3nI2.F1.ci}{}
\setres{cm.V3nI2.F1.n}{0}
\setres{cm.V3nI2.F2}{4}
\setres{cm.V3nI2.F2.ci}{[0, 10]}
\setres{cm.V3nI2.F2.n}{9}
\setres{cm.V3nI2.F3}{n/a}
\setres{cm.V3nI2.F3.ci}{}
\setres{cm.V3nI2.F3.n}{0}
\setres{cm.V3nI2.F4}{0}
\setres{cm.V3nI2.F4.ci}{[0, 0]}
\setres{cm.V3nI2.F4.n}{2}
\setres{cm.V3nI2.F5}{n/a}
\setres{cm.V3nI2.F5.ci}{}
\setres{cm.V3nI2.F5.n}{0}
\setres{cm.V3nI2.F6}{n/a}
\setres{cm.V3nI2.F6.ci}{}
\setres{cm.V3nI2.F6.n}{0}
\setres{cm.V3nI2.F7}{n/a}
\setres{cm.V3nI2.F7.ci}{}
\setres{cm.V3nI2.F7.n}{0}
\setres{cm.V3nI2.F8}{0}
\setres{cm.V3nI2.F8.ci}{[0, 0]}
\setres{cm.V3nI2.F8.n}{1}
\setres{cm.V3nI2.ci}{[0, 8]}
\setres{conf.gemma-3-27b.incons}{40}
\setres{conf.gemma-3-27b.safehigh}{96}
\setres{conf.gpt-oss-120b.incons}{30}
\setres{conf.gpt-oss-120b.safehigh}{100}
\setres{conf.gpt-oss-20b.incons}{51}
\setres{conf.gpt-oss-20b.safehigh}{100}
\setres{conf.incons.max}{51}
\setres{conf.incons.min}{27}
\setres{conf.medgemma-27b.incons}{45}
\setres{conf.medgemma-27b.safehigh}{99}
\setres{conf.qwen3-32b.incons}{27}
\setres{conf.qwen3-32b.safehigh}{73}
\setres{conf.qwen3-8b.incons}{32}
\setres{conf.qwen3-8b.safehigh}{88}
\setres{conf.safehigh.max}{100}
\setres{conf.safehigh.min}{73}
\setres{d.V1I1.F1}{0.07}
\setres{d.V1I1.F1.ci}{[0.01, 0.13]}
\setres{d.V1I1.F2}{$-$0.05}
\setres{d.V1I1.F2.ci}{[$-$0.08, $-$0.01]}
\setres{d.V1I1.F3}{$-$0.02}
\setres{d.V1I1.F3.ci}{[$-$0.05, 0.01]}
\setres{d.V1I1.F4}{$-$0.02}
\setres{d.V1I1.F4.ci}{[$-$0.04, 0.00]}
\setres{d.V1I1.F5}{$-$0.04}
\setres{d.V1I1.F5.ci}{[$-$0.06, 0.00]}
\setres{d.V1I1.F6}{0.00}
\setres{d.V1I1.F6.ci}{[$-$0.04, 0.04]}
\setres{d.V1I1.F7}{$-$0.04}
\setres{d.V1I1.F7.ci}{[$-$0.12, 0.03]}
\setres{d.V1I1.F8}{0.00}
\setres{d.V1I1.F8.ci}{[$-$0.03, 0.03]}
\setres{d.V1I1.all}{$-$0.01}
\setres{d.V1I1.all.ci}{[$-$0.03, 0.01]}
\setres{d.V1I1.all.ci.alt}{[$-$0.03, 0.01]}
\setres{d.V1I1.all.orig}{0.00}
\setres{d.V1I1.all.orig.ci}{[$-$0.02, 0.02]}
\setres{d.V1I1.ch}{0.01}
\setres{d.V1I1.ch.ci}{[$-$0.05, 0.06]}
\setres{d.V1I1.unch}{$-$0.01}
\setres{d.V1I1.unch.ci}{[$-$0.03, 0.00]}
\setres{d.V1I2.all}{$-$0.01}
\setres{d.V1I2.all.ci}{[$-$0.03, 0.01]}
\setres{d.V1I2.all.ci.alt}{[$-$0.03, 0.01]}
\setres{d.V1I2.all.orig}{$-$0.01}
\setres{d.V1I2.all.orig.ci}{[$-$0.04, 0.01]}
\setres{d.V1I2.ch}{0.00}
\setres{d.V1I2.ch.ci}{[$-$0.06, 0.05]}
\setres{d.V1I2.unch}{$-$0.01}
\setres{d.V1I2.unch.ci}{[$-$0.03, 0.01]}
\setres{d.V2I1.all}{0.39}
\setres{d.V2I1.all.ci}{[0.33, 0.45]}
\setres{d.V2I1.all.ci.alt}{[0.34, 0.45]}
\setres{d.V2I1.all.orig}{0.43}
\setres{d.V2I1.all.orig.ci}{[0.36, 0.50]}
\setres{d.V2I1.ch}{0.10}
\setres{d.V2I1.ch.ci}{[0.05, 0.15]}
\setres{d.V2I1.unch}{0.49}
\setres{d.V2I1.unch.ci}{[0.41, 0.57]}
\setres{d.V2I2.F1}{0.58}
\setres{d.V2I2.F1.ci}{[0.51, 0.65]}
\setres{d.V2I2.F2}{0.08}
\setres{d.V2I2.F2.ci}{[0.03, 0.14]}
\setres{d.V2I2.F3}{0.56}
\setres{d.V2I2.F3.ci}{[0.47, 0.63]}
\setres{d.V2I2.F4}{0.30}
\setres{d.V2I2.F4.ci}{[0.17, 0.43]}
\setres{d.V2I2.F5}{0.13}
\setres{d.V2I2.F5.ci}{[0.08, 0.18]}
\setres{d.V2I2.F6}{$-$0.01}
\setres{d.V2I2.F6.ci}{[$-$0.06, 0.04]}
\setres{d.V2I2.F7}{0.50}
\setres{d.V2I2.F7.ci}{[0.38, 0.61]}
\setres{d.V2I2.F8}{0.63}
\setres{d.V2I2.F8.ci}{[0.51, 0.74]}
\setres{d.V2I2.all}{0.38}
\setres{d.V2I2.all.ci}{[0.32, 0.45]}
\setres{d.V2I2.all.ci.alt}{[0.33, 0.44]}
\setres{d.V2I2.all.orig}{0.43}
\setres{d.V2I2.all.orig.ci}{[0.36, 0.50]}
\setres{d.V2I2.ch}{0.16}
\setres{d.V2I2.ch.ci}{[0.11, 0.21]}
\setres{d.V2I2.unch}{0.46}
\setres{d.V2I2.unch.ci}{[0.38, 0.54]}
\setres{d.V3I1.all}{0.77}
\setres{d.V3I1.all.ci}{[0.72, 0.82]}
\setres{d.V3I1.all.ci.alt}{[0.72, 0.82]}
\setres{d.V3I1.all.orig}{0.75}
\setres{d.V3I1.all.orig.ci}{[0.69, 0.81]}
\setres{d.V3I1.ch}{0.69}
\setres{d.V3I1.ch.ci}{[0.64, 0.74]}
\setres{d.V3I1.unch}{0.80}
\setres{d.V3I1.unch.ci}{[0.74, 0.86]}
\setres{d.V3I2.F1}{0.92}
\setres{d.V3I2.F1.ci}{[0.87, 0.96]}
\setres{d.V3I2.F2}{0.85}
\setres{d.V3I2.F2.ci}{[0.76, 0.93]}
\setres{d.V3I2.F3}{0.93}
\setres{d.V3I2.F3.ci}{[0.91, 0.96]}
\setres{d.V3I2.F4}{0.66}
\setres{d.V3I2.F4.ci}{[0.53, 0.77]}
\setres{d.V3I2.F5}{0.74}
\setres{d.V3I2.F5.ci}{[0.65, 0.84]}
\setres{d.V3I2.F6}{0.45}
\setres{d.V3I2.F6.ci}{[0.39, 0.51]}
\setres{d.V3I2.F7}{0.67}
\setres{d.V3I2.F7.ci}{[0.58, 0.76]}
\setres{d.V3I2.F8}{0.83}
\setres{d.V3I2.F8.ci}{[0.75, 0.90]}
\setres{d.V3I2.S012}{0.74}
\setres{d.V3I2.S012.ci}{[0.69, 0.79]}
\setres{d.V3I2.S3}{0.90}
\setres{d.V3I2.S3.ci}{[0.83, 0.96]}
\setres{d.V3I2.all}{0.75}
\setres{d.V3I2.all.ci}{[0.70, 0.80]}
\setres{d.V3I2.all.ci.alt}{[0.70, 0.79]}
\setres{d.V3I2.all.noreas}{0.67}
\setres{d.V3I2.all.noreas.ci}{[0.57, 0.75]}
\setres{d.V3I2.all.orig}{0.72}
\setres{d.V3I2.all.orig.ci}{[0.67, 0.78]}
\setres{d.V3I2.all.reas}{0.81}
\setres{d.V3I2.all.reas.ci}{[0.77, 0.85]}
\setres{d.V3I2.ch}{0.59}
\setres{d.V3I2.ch.ci}{[0.53, 0.64]}
\setres{d.V3I2.unch}{0.81}
\setres{d.V3I2.unch.ci}{[0.75, 0.86]}
\setres{d.V3I3.all}{0.78}
\setres{d.V3I3.all.ci}{[0.74, 0.82]}
\setres{d.V3I3.all.ci.alt}{[0.74, 0.82]}
\setres{d.V3I3.all.orig}{0.76}
\setres{d.V3I3.all.orig.ci}{[0.71, 0.81]}
\setres{d.V3I3.ch}{0.65}
\setres{d.V3I3.ch.ci}{[0.60, 0.70]}
\setres{d.V3I3.unch}{0.83}
\setres{d.V3I3.unch.ci}{[0.78, 0.88]}
\setres{d.V3nI2.all}{0.78}
\setres{d.V3nI2.all.ci}{[0.73, 0.82]}
\setres{d.V3nI2.all.ci.alt}{[0.73, 0.82]}
\setres{d.V3nI2.all.orig}{0.75}
\setres{d.V3nI2.all.orig.ci}{[0.70, 0.81]}
\setres{d.V3nI2.ch}{0.68}
\setres{d.V3nI2.ch.ci}{[0.62, 0.73]}
\setres{d.V3nI2.unch}{0.81}
\setres{d.V3nI2.unch.ci}{[0.75, 0.86]}
\setres{decoding.note}{Judges decode at temperature 0 with a fixed seed, except Qwen3 8B and Qwen3 32B, whose reasoning mode loops under greedy decoding and which use the vendor-recommended sampling (temperature 0.6, top-$p$ 0.95, top-$k$ 20) with the same seed. The gpt-oss models run with their native MXFP4 weights; all other models run in bf16.}
\setres{ece.V3I2}{0.26}
\setres{ece.gemma-3-27b.side}{0.09}
\setres{ece.gpt-oss-120b.side}{0.03}
\setres{ece.gpt-oss-20b.side}{0.19}
\setres{ece.medgemma-27b.side}{0.15}
\setres{ece.qwen3-32b.side}{0.07}
\setres{ece.qwen3-8b.side}{0.10}
\setres{ece.side.max}{0.19}
\setres{ece.side.min}{0.03}
\setres{ece.stated.max}{0.31}
\setres{ece.stated.min}{0.17}
\setres{f1.V3I2}{0.76}
\setres{fa.V1I1}{56}
\setres{fa.V1I1.ci}{[51, 61]}
\setres{fa.V1I2}{54}
\setres{fa.V1I2.ci}{[49, 59]}
\setres{fa.V2I1}{44}
\setres{fa.V2I1.ci}{[39, 50]}
\setres{fa.V2I2}{39}
\setres{fa.V2I2.ci}{[34, 44]}
\setres{fa.V3I1}{8}
\setres{fa.V3I1.ci}{[6, 11]}
\setres{fa.V3I2}{6}
\setres{fa.V3I2.ci}{[4, 9]}
\setres{fa.V3I3}{6}
\setres{fa.V3I3.ci}{[4, 8]}
\setres{fa.V3nI2}{7}
\setres{fa.V3nI2.ci}{[5, 9]}
\setres{fault.best.d}{0.93}
\setres{fault.best.id}{F3}
\setres{fault.best.name}{unit swap}
\setres{fault.worst.d}{0.45}
\setres{fault.worst.id}{F6}
\setres{fault.worst.name}{hallucinated observation}
\setres{joint.V1I1.S23.all}{23}
\setres{joint.V1I1.S23.any}{158}
\setres{joint.V1I1.S23.k}{6}
\setres{joint.V1I1.S23.maj}{48}
\setres{joint.V1I1.S23.n}{223}
\setres{joint.V1I1.S3.all}{0}
\setres{joint.V1I1.S3.any}{5}
\setres{joint.V1I1.S3.k}{6}
\setres{joint.V1I1.S3.maj}{0}
\setres{joint.V1I1.S3.n}{12}
\setres{joint.V1I2.S23.all}{14}
\setres{joint.V1I2.S23.any}{199}
\setres{joint.V1I2.S23.k}{6}
\setres{joint.V1I2.S23.maj}{50}
\setres{joint.V1I2.S23.n}{223}
\setres{joint.V1I2.S3.all}{0}
\setres{joint.V1I2.S3.any}{9}
\setres{joint.V1I2.S3.k}{6}
\setres{joint.V1I2.S3.maj}{0}
\setres{joint.V1I2.S3.n}{12}
\setres{joint.V2I1.S23.all}{6}
\setres{joint.V2I1.S23.any}{58}
\setres{joint.V2I1.S23.k}{6}
\setres{joint.V2I1.S23.maj}{22}
\setres{joint.V2I1.S23.n}{222}
\setres{joint.V2I1.S3.all}{0}
\setres{joint.V2I1.S3.any}{0}
\setres{joint.V2I1.S3.k}{6}
\setres{joint.V2I1.S3.maj}{0}
\setres{joint.V2I1.S3.n}{12}
\setres{joint.V2I2.S23.all}{9}
\setres{joint.V2I2.S23.any}{82}
\setres{joint.V2I2.S23.k}{6}
\setres{joint.V2I2.S23.maj}{27}
\setres{joint.V2I2.S23.n}{222}
\setres{joint.V2I2.S3.all}{0}
\setres{joint.V2I2.S3.any}{0}
\setres{joint.V2I2.S3.k}{6}
\setres{joint.V2I2.S3.maj}{0}
\setres{joint.V2I2.S3.n}{12}
\setres{joint.V3I1.S23.all}{1}
\setres{joint.V3I1.S23.any}{66}
\setres{joint.V3I1.S23.k}{6}
\setres{joint.V3I1.S23.maj}{9}
\setres{joint.V3I1.S23.n}{223}
\setres{joint.V3I1.S3.all}{0}
\setres{joint.V3I1.S3.any}{6}
\setres{joint.V3I1.S3.k}{6}
\setres{joint.V3I1.S3.maj}{0}
\setres{joint.V3I1.S3.n}{12}
\setres{joint.V3I2.S23.all}{0}
\setres{joint.V3I2.S23.any}{84}
\setres{joint.V3I2.S23.k}{6}
\setres{joint.V3I2.S23.maj}{18}
\setres{joint.V3I2.S23.n}{223}
\setres{joint.V3I2.S3.all}{0}
\setres{joint.V3I2.S3.any}{4}
\setres{joint.V3I2.S3.k}{6}
\setres{joint.V3I2.S3.maj}{0}
\setres{joint.V3I2.S3.n}{12}
\setres{joint.V3I3.S23.all}{0}
\setres{joint.V3I3.S23.any}{75}
\setres{joint.V3I3.S23.k}{6}
\setres{joint.V3I3.S23.maj}{14}
\setres{joint.V3I3.S23.n}{223}
\setres{joint.V3I3.S3.all}{0}
\setres{joint.V3I3.S3.any}{2}
\setres{joint.V3I3.S3.k}{6}
\setres{joint.V3I3.S3.maj}{0}
\setres{joint.V3I3.S3.n}{12}
\setres{joint.V3nI2.S23.all}{0}
\setres{joint.V3nI2.S23.any}{64}
\setres{joint.V3nI2.S23.k}{6}
\setres{joint.V3nI2.S23.maj}{5}
\setres{joint.V3nI2.S23.n}{222}
\setres{joint.V3nI2.S3.all}{0}
\setres{joint.V3nI2.S3.any}{2}
\setres{joint.V3nI2.S3.k}{6}
\setres{joint.V3nI2.S3.maj}{0}
\setres{joint.V3nI2.S3.n}{11}
\setres{judge.best.cm}{0}
\setres{judge.best.d}{0.81}
\setres{judge.best.fa}{3}
\setres{judge.best.name}{Gemma 3 27B}
\setres{judge.best.ppv1}{25}
\setres{judge.best.ppv5}{63}
\setres{judge.gemma-3-27b.cm}{0}
\setres{judge.gemma-3-27b.cm.ci}{[0, 0]}
\setres{judge.gemma-3-27b.fa}{3}
\setres{judge.gemma-3-27b.fa.ci}{[1, 5]}
\setres{judge.gemma-3-27b.sens}{85}
\setres{judge.gemma-3-27b.sens.ci}{[81, 88]}
\setres{judge.gemma-3-27b.sm}{17}
\setres{judge.gemma-3-27b.sm.ci}{[12, 22]}
\setres{judge.gpt-oss-120b.cm}{0}
\setres{judge.gpt-oss-120b.cm.ci}{[0, 0]}
\setres{judge.gpt-oss-120b.fa}{14}
\setres{judge.gpt-oss-120b.fa.ci}{[9, 19]}
\setres{judge.gpt-oss-120b.sens}{96}
\setres{judge.gpt-oss-120b.sens.ci}{[94, 98]}
\setres{judge.gpt-oss-120b.sm}{0}
\setres{judge.gpt-oss-120b.sm.ci}{[0, 0]}
\setres{judge.gpt-oss-20b.cm}{0}
\setres{judge.gpt-oss-20b.cm.ci}{[0, 0]}
\setres{judge.gpt-oss-20b.fa}{1}
\setres{judge.gpt-oss-20b.fa.ci}{[0, 3]}
\setres{judge.gpt-oss-20b.sens}{71}
\setres{judge.gpt-oss-20b.sens.ci}{[66, 76]}
\setres{judge.gpt-oss-20b.sm}{21}
\setres{judge.gpt-oss-20b.sm.ci}{[15, 27]}
\setres{judge.medgemma-27b.cm}{33}
\setres{judge.medgemma-27b.cm.ci}{[8, 67]}
\setres{judge.medgemma-27b.fa}{3}
\setres{judge.medgemma-27b.fa.ci}{[1, 5]}
\setres{judge.medgemma-27b.sens}{78}
\setres{judge.medgemma-27b.sens.ci}{[74, 82]}
\setres{judge.medgemma-27b.sm}{21}
\setres{judge.medgemma-27b.sm.ci}{[16, 26]}
\setres{judge.qwen3-32b.cm}{0}
\setres{judge.qwen3-32b.cm.ci}{[0, 0]}
\setres{judge.qwen3-32b.fa}{5}
\setres{judge.qwen3-32b.fa.ci}{[2, 9]}
\setres{judge.qwen3-32b.sens}{89}
\setres{judge.qwen3-32b.sens.ci}{[86, 92]}
\setres{judge.qwen3-32b.sm}{13}
\setres{judge.qwen3-32b.sm.ci}{[8, 17]}
\setres{judge.qwen3-8b.cm}{0}
\setres{judge.qwen3-8b.cm.ci}{[0, 0]}
\setres{judge.qwen3-8b.fa}{13}
\setres{judge.qwen3-8b.fa.ci}{[8, 19]}
\setres{judge.qwen3-8b.sens}{87}
\setres{judge.qwen3-8b.sens.ci}{[84, 90]}
\setres{judge.qwen3-8b.sm}{14}
\setres{judge.qwen3-8b.sm.ci}{[10, 20]}
\setres{judge.worst.d}{0.67}
\setres{judge.worst.name}{Qwen3 8B}
\setres{judges.list}{Qwen3 8B, Qwen3 32B, gpt-oss 20B, gpt-oss 120B, Gemma 3 27B, and MedGemma 27B}
\setres{loc.V3I2}{100}
\setres{lucky.confirmed}{0}
\setres{lucky.n}{9}
\setres{lucky.rate}{1.5}
\setres{lucky.reviewed}{0}
\setres{lucky.top.flag}{deciding not to order without reading the deciding lab}
\setres{n.S23}{223}
\setres{n.S3}{12}
\setres{n.V3I2.noreas}{1,062}
\setres{n.V3I2.reas}{1,434}
\setres{n.agent.models}{2}
\setres{n.clusters}{162}
\setres{n.conds}{8}
\setres{n.controls}{60}
\setres{n.episodes}{1,260}
\setres{n.examples}{12}
\setres{n.hyp}{6}
\setres{n.items}{609}
\setres{n.items.all}{669}
\setres{n.judges}{6}
\setres{n.missed.cases}{0}
\setres{n.mutants}{416}
\setres{n.parents}{193}
\setres{n.parents.agent}{137}
\setres{n.parents.oracle}{56}
\setres{n.passed}{587}
\setres{n.per.fault}{18}
\setres{n.runs.per.task}{2}
\setres{n.tasks}{315}
\setres{n.tasks.ext}{15}
\setres{n.tasks.in.items}{162}
\setres{n.tasks.mutants}{162}
\setres{n.verdicts}{29,232}
\setres{n.verdicts.all}{32,112}
\setres{nat.added.clean}{0}
\setres{nat.added.clean.n}{0}
\setres{nat.added.clean.tasks}{0}
\setres{nat.added.pass}{0}
\setres{nat.added.runs}{60}
\setres{nat.added.tasks}{15}
\setres{nat.anyflag}{25}
\setres{nat.claim.all}{7}
\setres{nat.claim.failed}{14}
\setres{nat.claim.n}{94}
\setres{nat.claim.passed.n}{0}
\setres{nat.dosing.all}{1}
\setres{nat.dosing.failed}{1}
\setres{nat.dosing.n}{10}
\setres{nat.dosing.passed.n}{0}
\setres{nat.failed}{673}
\setres{nat.failed.clean.n}{368}
\setres{nat.failed.flagged}{45}
\setres{nat.invalidpost.all}{4}
\setres{nat.invalidpost.failed}{7}
\setres{nat.invalidpost.n}{50}
\setres{nat.invalidpost.passed.n}{0}
\setres{nat.nolab.all}{2}
\setres{nat.nolab.failed}{3}
\setres{nat.nolab.n}{22}
\setres{nat.nolab.passed.n}{0}
\setres{nat.nolabread.all}{12}
\setres{nat.nolabread.failed}{20}
\setres{nat.nolabread.n}{146}
\setres{nat.nolabread.passed.n}{9}
\setres{nat.order.clean}{24}
\setres{nat.order.clean.n}{58}
\setres{nat.order.clean.tasks}{18}
\setres{nat.order.pass}{28}
\setres{nat.order.runs}{240}
\setres{nat.order.tasks}{60}
\setres{nat.other.clean}{54}
\setres{nat.other.clean.n}{520}
\setres{nat.other.clean.tasks}{197}
\setres{nat.other.pass}{54}
\setres{nat.other.runs}{960}
\setres{nat.other.tasks}{240}
\setres{nat.posts.all}{10}
\setres{nat.posts.failed}{18}
\setres{nat.posts.n}{121}
\setres{nat.posts.passed.n}{0}
\setres{nat.runs}{1,260}
\setres{nat.status.all}{5}
\setres{nat.status.failed}{10}
\setres{nat.status.n}{67}
\setres{nat.status.passed.n}{0}
\setres{nat.unchecked.all}{1}
\setres{nat.unchecked.failed}{1}
\setres{nat.unchecked.n}{7}
\setres{nat.unchecked.passed.n}{0}
\setres{nat.ungrounded.all}{0}
\setres{nat.ungrounded.failed}{1}
\setres{nat.ungrounded.n}{5}
\setres{nat.ungrounded.passed.n}{0}
\setres{panel.best.single}{0.81}
\setres{panel.d}{0.85}
\setres{panel.k}{6}
\setres{panel.neff}{2.2}
\setres{parse.fail}{0.0}
\setres{pass.rate}{47}
\setres{profile}{fast}
\setres{reasoning.F1}{0.01}
\setres{reasoning.F1.ci}{[$-$0.02, 0.04]}
\setres{reasoning.F2}{$-$0.04}
\setres{reasoning.F2.ci}{[$-$0.11, 0.02]}
\setres{reasoning.F3}{0.01}
\setres{reasoning.F3.ci}{[$-$0.03, 0.04]}
\setres{reasoning.F4}{0.00}
\setres{reasoning.F4.ci}{[$-$0.04, 0.04]}
\setres{reasoning.F5}{$-$0.02}
\setres{reasoning.F5.ci}{[$-$0.07, 0.03]}
\setres{reasoning.F6}{$-$0.18}
\setres{reasoning.F6.ci}{[$-$0.24, $-$0.12]}
\setres{reasoning.F7}{0.03}
\setres{reasoning.F7.ci}{[$-$0.02, 0.08]}
\setres{reasoning.F8}{$-$0.03}
\setres{reasoning.F8.ci}{[$-$0.06, 0.00]}
\setres{reasoning.all}{$-$0.03}
\setres{reasoning.all.ci}{[$-$0.05, $-$0.01]}
\setres{reasoning.ch}{$-$0.10}
\setres{reasoning.ch.ci}{[$-$0.14, $-$0.06]}
\setres{reasoning.unch}{0.00}
\setres{reasoning.unch.ci}{[$-$0.02, 0.02]}
\setres{recheck.flagged.before}{28}
\setres{recheck.flagged.now}{9}
\setres{recheck.items.dropped}{0}
\setres{recheck.rejected}{0}
\setres{responses.F1}{0.33}
\setres{responses.F1.ci}{[0.28, 0.38]}
\setres{responses.F2}{0.81}
\setres{responses.F2.ci}{[0.72, 0.90]}
\setres{responses.F3}{0.37}
\setres{responses.F3.ci}{[0.29, 0.47]}
\setres{responses.F4}{0.35}
\setres{responses.F4.ci}{[0.23, 0.48]}
\setres{responses.F5}{0.63}
\setres{responses.F5.ci}{[0.55, 0.71]}
\setres{responses.F6}{0.64}
\setres{responses.F6.ci}{[0.57, 0.71]}
\setres{responses.F7}{0.15}
\setres{responses.F7.ci}{[0.10, 0.21]}
\setres{responses.F8}{0.23}
\setres{responses.F8.ci}{[0.18, 0.29]}
\setres{responses.all}{0.39}
\setres{responses.all.ci}{[0.34, 0.44]}
\setres{responses.ch}{0.52}
\setres{responses.ch.ci}{[0.46, 0.58]}
\setres{responses.unch}{0.35}
\setres{responses.unch.ci}{[0.28, 0.41]}
\setres{retest.agree}{93.4}
\setres{retest.kappa}{0.85}
\setres{sens.V1I1}{61}
\setres{sens.V1I1.ci}{[57, 66]}
\setres{sens.V1I2}{56}
\setres{sens.V1I2.ci}{[51, 61]}
\setres{sens.V2I1}{91}
\setres{sens.V2I1.ci}{[88, 93]}
\setres{sens.V2I2}{87}
\setres{sens.V2I2.ci}{[84, 89]}
\setres{sens.V3I1}{89}
\setres{sens.V3I1.ci}{[87, 91]}
\setres{sens.V3I2}{84}
\setres{sens.V3I2.ci}{[82, 87]}
\setres{sens.V3I3}{87}
\setres{sens.V3I3.ci}{[85, 89]}
\setres{sens.V3nI2}{87}
\setres{sens.V3nI2.ci}{[85, 89]}
\setres{sent.H6}{MedGemma 27B differs from Gemma 3 27B, the general model of the same size, by $-$0.08 [$-$0.14, $-$0.02] (H6, $p$\,$<$\,0.001).}
\setres{sent.H7}{Judges' sensitivity on mutants of trajectories from their own model family differs from that on other families by $-$0.10 [$-$0.15, $-$0.05] (H7, exploratory, unadjusted $p$\,$<$\,0.001).}
\setres{sent.controls}{On the 60 harmless-edit controls, judges under V3--I2 flagged the edited copy in 7\% of verdicts and its unedited parent in 7\% (difference 0.00 [$-$0.02, 0.02]).}
\setres{sent.lucky}{These flags are rule-based and were not manually adjudicated.}
\setres{sent.lucky.main}{The strict checker flagged 1.5\% of the grader-passed runs (9 of 587), most often for deciding not to order without reading the deciding lab.}
\setres{sent.panel}{A majority vote of all 6 judges reaches $\Delta = 0.85$, against 0.81 for the best single judge; by error correlation the panel carries $n_{\text{eff}} = 2.2$ independent votes.}
\setres{sent.reasoning}{Showing the EHR responses without the agent's reasoning (V3n) changes pooled $\Delta$ by 0.39 [0.34, 0.44] over the action-only view; adding the agent's recorded reasoning (V3) changes it by a further $-$0.03 [$-$0.05, $-$0.01].}
\setres{sent.recheck}{Every selected parent also passes the final checker rules.}
\setres{sent.robust}{Intervals resample tasks (162 clusters). Resampling parent trajectories instead gives [0.70, 0.79] for the pooled V3--I2 $\Delta$, and on the original MedAgentBench tasks alone, without the added medication-order tasks, it is 0.72 [0.67, 0.78].}
\setres{sent.severe}{With every action and EHR response in view (V3--I2), judges missed 6\% [1, 11] of the 12 faults rated S3, pooled over judges, while flagging 6\% [4, 9] of clean parents. Of the 12 S3 faults that all 6 judges rated, 0 were missed by every judge and 0 by a majority.}
\setres{sent.severe.scope}{The 12 S3 faults come from 11 tasks of 2 task types (9 dose magnitude, 2 ignored observation and 1 wrong code); 83\% are in the added medication-order tasks and 100\% leave the final answer unchanged.}
\setres{sent.spotcheck}{}
\setres{share.ext}{14}
\setres{share.grader.silent}{9}
\setres{share.unchanged}{74}
\setres{sm.V1I1}{32}
\setres{sm.V1I1.ci}{[26, 39]}
\setres{sm.V1I2}{37}
\setres{sm.V1I2.ci}{[32, 42]}
\setres{sm.V2I1}{13}
\setres{sm.V2I1.ci}{[10, 17]}
\setres{sm.V2I2}{16}
\setres{sm.V2I2.ci}{[13, 20]}
\setres{sm.V3I1}{10}
\setres{sm.V3I1.ci}{[7, 12]}
\setres{sm.V3I2}{14}
\setres{sm.V3I2.ci}{[11, 18]}
\setres{sm.V3I3}{12}
\setres{sm.V3I3.ci}{[9, 15]}
\setres{sm.V3nI2}{9}
\setres{sm.V3nI2.ci}{[7, 12]}
\setres{spotcheck.rejected}{0}
\setres{spotcheck.reviewed}{0}
\setres{spotcheck.sampled}{56}
\setres{stack.ALL.anyjudge}{98}
\setres{stack.ALL.anyjudge.ci}{[97, 99]}
\setres{stack.ALL.checker}{71}
\setres{stack.ALL.checker.ci}{[67, 75]}
\setres{stack.ALL.cj}{97}
\setres{stack.ALL.cj.ci}{[95, 98]}
\setres{stack.ALL.either}{100}
\setres{stack.ALL.either.ci}{[100, 100]}
\setres{stack.ALL.grader}{91}
\setres{stack.ALL.grader.ci}{[89, 93]}
\setres{stack.ALL.judges}{84}
\setres{stack.ALL.judges.ci}{[82, 87]}
\setres{stack.ALL.maj}{87}
\setres{stack.ALL.maj.ci}{[83, 90]}
\setres{stack.ALL.majmiss.checker}{42}
\setres{stack.ALL.majmiss.either}{56}
\setres{stack.ALL.majmiss.grader}{54}
\setres{stack.ALL.majmiss.n}{56}
\setres{stack.ALL.n}{416}
\setres{stack.ALL.none}{0}
\setres{stack.ALL.none.ci}{[0, 0]}
\setres{stack.ALL.none.n}{0}
\setres{stack.F1.anyjudge}{100}
\setres{stack.F1.anyjudge.ci}{[100, 100]}
\setres{stack.F1.checker}{100}
\setres{stack.F1.checker.ci}{[100, 100]}
\setres{stack.F1.cj}{100}
\setres{stack.F1.cj.ci}{[100, 100]}
\setres{stack.F1.either}{100}
\setres{stack.F1.either.ci}{[100, 100]}
\setres{stack.F1.grader}{100}
\setres{stack.F1.grader.ci}{[100, 100]}
\setres{stack.F1.judges}{98}
\setres{stack.F1.judges.ci}{[96, 99]}
\setres{stack.F1.maj}{100}
\setres{stack.F1.maj.ci}{[100, 100]}
\setres{stack.F1.majmiss.checker}{0}
\setres{stack.F1.majmiss.either}{0}
\setres{stack.F1.majmiss.grader}{0}
\setres{stack.F1.majmiss.n}{0}
\setres{stack.F1.n}{60}
\setres{stack.F1.none}{0}
\setres{stack.F1.none.ci}{[0, 0]}
\setres{stack.F1.none.n}{0}
\setres{stack.F2.anyjudge}{100}
\setres{stack.F2.anyjudge.ci}{[100, 100]}
\setres{stack.F2.checker}{100}
\setres{stack.F2.checker.ci}{[100, 100]}
\setres{stack.F2.cj}{100}
\setres{stack.F2.cj.ci}{[100, 100]}
\setres{stack.F2.either}{100}
\setres{stack.F2.either.ci}{[100, 100]}
\setres{stack.F2.grader}{100}
\setres{stack.F2.grader.ci}{[100, 100]}
\setres{stack.F2.judges}{93}
\setres{stack.F2.judges.ci}{[87, 98]}
\setres{stack.F2.maj}{100}
\setres{stack.F2.maj.ci}{[100, 100]}
\setres{stack.F2.majmiss.checker}{0}
\setres{stack.F2.majmiss.either}{0}
\setres{stack.F2.majmiss.grader}{0}
\setres{stack.F2.majmiss.n}{0}
\setres{stack.F2.n}{18}
\setres{stack.F2.none}{0}
\setres{stack.F2.none.ci}{[0, 0]}
\setres{stack.F2.none.n}{0}
\setres{stack.F3.anyjudge}{100}
\setres{stack.F3.anyjudge.ci}{[100, 100]}
\setres{stack.F3.checker}{30}
\setres{stack.F3.checker.ci}{[18, 44]}
\setres{stack.F3.cj}{100}
\setres{stack.F3.cj.ci}{[100, 100]}
\setres{stack.F3.either}{100}
\setres{stack.F3.either.ci}{[100, 100]}
\setres{stack.F3.grader}{100}
\setres{stack.F3.grader.ci}{[100, 100]}
\setres{stack.F3.judges}{99}
\setres{stack.F3.judges.ci}{[97, 100]}
\setres{stack.F3.maj}{100}
\setres{stack.F3.maj.ci}{[100, 100]}
\setres{stack.F3.majmiss.checker}{0}
\setres{stack.F3.majmiss.either}{0}
\setres{stack.F3.majmiss.grader}{0}
\setres{stack.F3.majmiss.n}{0}
\setres{stack.F3.n}{60}
\setres{stack.F3.none}{0}
\setres{stack.F3.none.ci}{[0, 0]}
\setres{stack.F3.none.n}{0}
\setres{stack.F4.anyjudge}{88}
\setres{stack.F4.anyjudge.ci}{[79, 96]}
\setres{stack.F4.checker}{100}
\setres{stack.F4.checker.ci}{[100, 100]}
\setres{stack.F4.cj}{100}
\setres{stack.F4.cj.ci}{[100, 100]}
\setres{stack.F4.either}{100}
\setres{stack.F4.either.ci}{[100, 100]}
\setres{stack.F4.grader}{100}
\setres{stack.F4.grader.ci}{[100, 100]}
\setres{stack.F4.judges}{78}
\setres{stack.F4.judges.ci}{[69, 85]}
\setres{stack.F4.maj}{85}
\setres{stack.F4.maj.ci}{[75, 94]}
\setres{stack.F4.majmiss.checker}{9}
\setres{stack.F4.majmiss.either}{9}
\setres{stack.F4.majmiss.grader}{9}
\setres{stack.F4.majmiss.n}{9}
\setres{stack.F4.n}{60}
\setres{stack.F4.none}{0}
\setres{stack.F4.none.ci}{[0, 0]}
\setres{stack.F4.none.n}{0}
\setres{stack.F5.anyjudge}{100}
\setres{stack.F5.anyjudge.ci}{[100, 100]}
\setres{stack.F5.checker}{100}
\setres{stack.F5.checker.ci}{[100, 100]}
\setres{stack.F5.cj}{100}
\setres{stack.F5.cj.ci}{[100, 100]}
\setres{stack.F5.either}{100}
\setres{stack.F5.either.ci}{[100, 100]}
\setres{stack.F5.grader}{0}
\setres{stack.F5.grader.ci}{[0, 0]}
\setres{stack.F5.judges}{95}
\setres{stack.F5.judges.ci}{[91, 99]}
\setres{stack.F5.maj}{95}
\setres{stack.F5.maj.ci}{[88, 100]}
\setres{stack.F5.majmiss.checker}{2}
\setres{stack.F5.majmiss.either}{2}
\setres{stack.F5.majmiss.grader}{0}
\setres{stack.F5.majmiss.n}{2}
\setres{stack.F5.n}{38}
\setres{stack.F5.none}{0}
\setres{stack.F5.none.ci}{[0, 0]}
\setres{stack.F5.none.n}{0}
\setres{stack.F6.anyjudge}{98}
\setres{stack.F6.anyjudge.ci}{[95, 100]}
\setres{stack.F6.checker}{70}
\setres{stack.F6.checker.ci}{[58, 81]}
\setres{stack.F6.cj}{77}
\setres{stack.F6.cj.ci}{[66, 87]}
\setres{stack.F6.either}{100}
\setres{stack.F6.either.ci}{[100, 100]}
\setres{stack.F6.grader}{100}
\setres{stack.F6.grader.ci}{[100, 100]}
\setres{stack.F6.judges}{49}
\setres{stack.F6.judges.ci}{[43, 55]}
\setres{stack.F6.maj}{35}
\setres{stack.F6.maj.ci}{[23, 47]}
\setres{stack.F6.majmiss.checker}{25}
\setres{stack.F6.majmiss.either}{39}
\setres{stack.F6.majmiss.grader}{39}
\setres{stack.F6.majmiss.n}{39}
\setres{stack.F6.n}{60}
\setres{stack.F6.none}{0}
\setres{stack.F6.none.ci}{[0, 0]}
\setres{stack.F6.none.n}{0}
\setres{stack.F7.anyjudge}{100}
\setres{stack.F7.anyjudge.ci}{[100, 100]}
\setres{stack.F7.checker}{100}
\setres{stack.F7.checker.ci}{[100, 100]}
\setres{stack.F7.cj}{100}
\setres{stack.F7.cj.ci}{[100, 100]}
\setres{stack.F7.either}{100}
\setres{stack.F7.either.ci}{[100, 100]}
\setres{stack.F7.grader}{100}
\setres{stack.F7.grader.ci}{[100, 100]}
\setres{stack.F7.judges}{80}
\setres{stack.F7.judges.ci}{[76, 85]}
\setres{stack.F7.maj}{90}
\setres{stack.F7.maj.ci}{[81, 98]}
\setres{stack.F7.majmiss.checker}{6}
\setres{stack.F7.majmiss.either}{6}
\setres{stack.F7.majmiss.grader}{6}
\setres{stack.F7.majmiss.n}{6}
\setres{stack.F7.n}{60}
\setres{stack.F7.none}{0}
\setres{stack.F7.none.ci}{[0, 0]}
\setres{stack.F7.none.n}{0}
\setres{stack.F8.anyjudge}{100}
\setres{stack.F8.anyjudge.ci}{[100, 100]}
\setres{stack.F8.checker}{0}
\setres{stack.F8.checker.ci}{[0, 0]}
\setres{stack.F8.cj}{100}
\setres{stack.F8.cj.ci}{[100, 100]}
\setres{stack.F8.either}{100}
\setres{stack.F8.either.ci}{[100, 100]}
\setres{stack.F8.grader}{100}
\setres{stack.F8.grader.ci}{[100, 100]}
\setres{stack.F8.judges}{95}
\setres{stack.F8.judges.ci}{[93, 97]}
\setres{stack.F8.maj}{100}
\setres{stack.F8.maj.ci}{[100, 100]}
\setres{stack.F8.majmiss.checker}{0}
\setres{stack.F8.majmiss.either}{0}
\setres{stack.F8.majmiss.grader}{0}
\setres{stack.F8.majmiss.n}{0}
\setres{stack.F8.n}{60}
\setres{stack.F8.none}{0}
\setres{stack.F8.none.ci}{[0, 0]}
\setres{stack.F8.none.n}{0}
\setres{stack.S23.anyjudge}{100}
\setres{stack.S23.anyjudge.ci}{[100, 100]}
\setres{stack.S23.checker}{78}
\setres{stack.S23.checker.ci}{[72, 84]}
\setres{stack.S23.cj}{97}
\setres{stack.S23.cj.ci}{[95, 99]}
\setres{stack.S23.either}{100}
\setres{stack.S23.either.ci}{[100, 100]}
\setres{stack.S23.grader}{92}
\setres{stack.S23.grader.ci}{[89, 95]}
\setres{stack.S23.judges}{86}
\setres{stack.S23.judges.ci}{[82, 89]}
\setres{stack.S23.maj}{86}
\setres{stack.S23.maj.ci}{[81, 91]}
\setres{stack.S23.majmiss.checker}{25}
\setres{stack.S23.majmiss.either}{31}
\setres{stack.S23.majmiss.grader}{31}
\setres{stack.S23.majmiss.n}{31}
\setres{stack.S23.n}{223}
\setres{stack.S23.none}{0}
\setres{stack.S23.none.ci}{[0, 0]}
\setres{stack.S23.none.n}{0}
\setres{stack.S3.anyjudge}{100}
\setres{stack.S3.anyjudge.ci}{[100, 100]}
\setres{stack.S3.checker}{92}
\setres{stack.S3.checker.ci}{[77, 100]}
\setres{stack.S3.cj}{100}
\setres{stack.S3.cj.ci}{[100, 100]}
\setres{stack.S3.either}{100}
\setres{stack.S3.either.ci}{[100, 100]}
\setres{stack.S3.grader}{100}
\setres{stack.S3.grader.ci}{[100, 100]}
\setres{stack.S3.judges}{94}
\setres{stack.S3.judges.ci}{[89, 99]}
\setres{stack.S3.maj}{100}
\setres{stack.S3.maj.ci}{[100, 100]}
\setres{stack.S3.majmiss.checker}{0}
\setres{stack.S3.majmiss.either}{0}
\setres{stack.S3.majmiss.grader}{0}
\setres{stack.S3.majmiss.n}{0}
\setres{stack.S3.n}{12}
\setres{stack.S3.none}{0}
\setres{stack.S3.none.ci}{[0, 0]}
\setres{stack.S3.none.n}{0}
\setres{stack.controls.flagged}{0}
\setres{stack.errors}{0}
\setres{stack.parents.flagged}{0}
\setres{truncated}{0.0}
\setres{vllm.version}{0.31.0}
}{}
\defrows{tab_view_rows}{5}\defrows{tab_severe_rows}{6}\defrows{tab_judges_rows}{8}\defrows{tab_agents_rows}{5}\defrows{tab_hyp_rows}{4}\defrows{tab_judges_side_rows}{8}\defrows{tab_stack_rows}{7}
\newcommand{\resfig}[2]{\IfFileExists{\resdir #1.pdf}{\includegraphics[width=#2]{\resdir #1}}{\fbox{\parbox[c][3cm][c]{0.95#2}{\centering\small\textcolor{red}{figure pending: \texttt{\detokenize{#1}.pdf}}}}}}

\newcommand{\clean}{\tau_0}
\newcommand{\mut}{\tau}
\newcommand{\view}{V}
\newcommand{\Dlt}{\Delta}

\begin{document}

\title{Can LLM Judges Catch Unsafe Clinical Actions? \\ A Fault-Injection Study}

\author{
\IEEEauthorblockN{Neda Ghafouri}
\IEEEauthorblockA{Computer Science Department}
\textit{University of Central Florida, Orlando, FL, USA}
\textit{neda.ghafouri@ucf.edu}
}

\maketitle









\begin{abstract}
As clinical AI agents begin to interact with electronic health records (EHRs), large language models (LLMs) are increasingly used to evaluate their actions. But how reliably can these judges recognize clinically dangerous mistakes, especially when an agent appears to have completed its task successfully?
We investigate this question using MedAgentBench by introducing controlled clinical errors into verified agent trajectories. The experiments include 416 faulty trajectories, eight error categories, and six open-weight LLM judges. By comparing each faulty trajectory with its correct counterpart, we measure whether judges genuinely recognize errors rather than simply flagging more actions as unsafe.
The findings show that access to clinical evidence significantly improves judge reliability. When shown only the final answer, judges cannot distinguish errors that leave it unchanged, while flagging 56\% of correct trajectories. With full trajectory information, the ability to distinguish unsafe actions from correct ones reaches a score of 0.75, and providing actual EHR responses improves discrimination by 0.39. Surprisingly, including the agent's reasoning does not improve overall performance and reduces detection of incorrectly reported clinical values by 0.18, which remain the hardest errors to detect. MedGemma performed worse than the general-purpose model and was the only judge to miss potentially dangerous medication overdoses. We also find that the benchmark's original grader accepts all 38 trajectories with an intentionally skipped laboratory check, while combining LLM judges with a rule-based checker informed by the full EHR record detects 97\% of injected faults.
\end{abstract}


\section{Introduction}
\label{sec:intro}
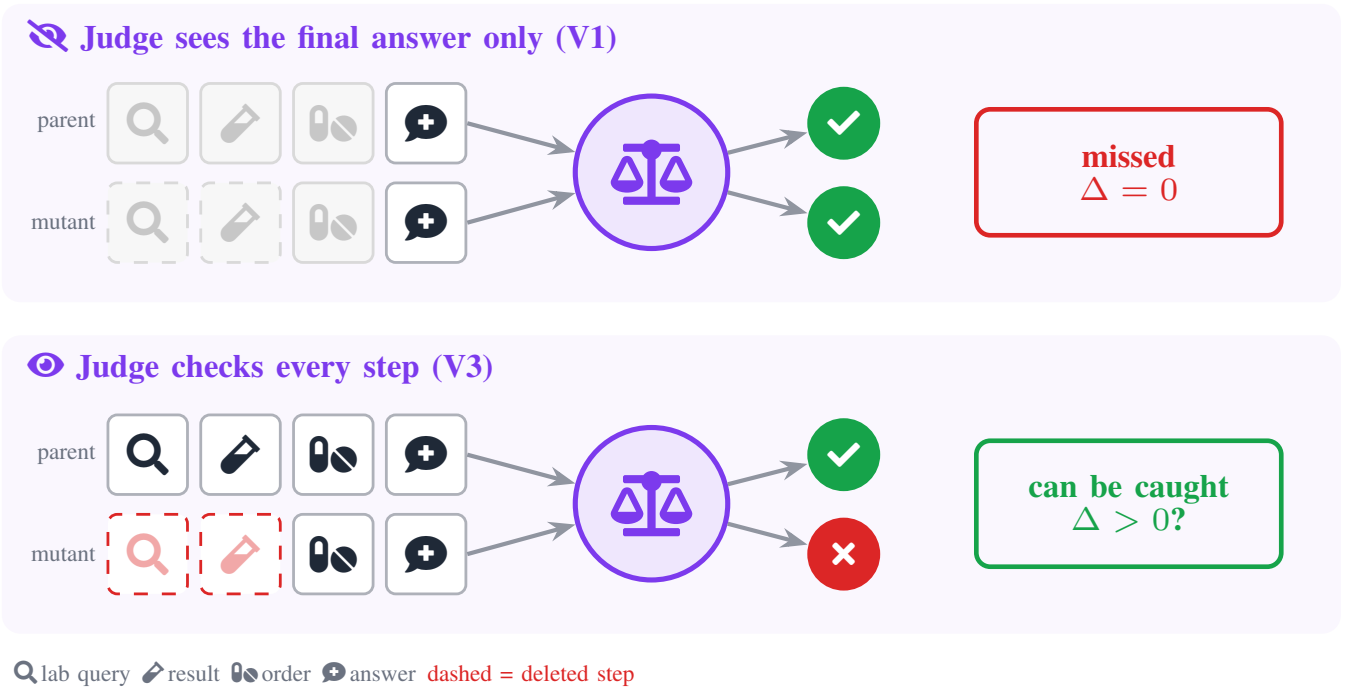
\begin{figure}[t]
\centering
\definecolor{cFault}{RGB}{220,38,38}
\definecolor{cJudge}{RGB}{124,58,237}
\definecolor{cOK}{RGB}{22,163,74}
\definecolor{cInk}{RGB}{31,41,55}
\definecolor{cMute}{RGB}{107,114,128}
\tikzset{
  av/.style={circle, fill=#1!12, draw=#1, line width=1.1pt, minimum size=1.15cm, text=#1, inner sep=0pt, font=\Large},
  lab/.style={font=\scriptsize\bfseries, text=cInk, align=center, inner sep=1pt},
  sub/.style={font=\tiny, text=cMute, align=center, inner sep=1pt},
  tile/.style={rounded corners=2pt, draw=cMute!55, fill=white, minimum width=0.6cm, minimum height=0.6cm, inner sep=0pt, text=cInk, line width=0.6pt, font=\small},
  tgone/.style={tile, draw=cFault, dashed, fill=white, text=cFault!40},
  tfade/.style={tile, fill=black!3, text=black!22, draw=black!15},
  tfadegone/.style={tile, fill=black!3, text=black!22, draw=black!15, dashed},
  flow/.style={-{Stealth[length=2mm]}, line width=0.9pt, draw=cMute!75},
  bigbadge/.style={circle, fill=#1, text=white, inner sep=0pt, minimum size=0.55cm, font=\scriptsize},
  pics/run/.style args={#1/#2/#3/#4}{code={
    \node[#1] (-t1) at (0,0)   {\faSearch};
    \node[#2] (-t2) at (0.7,0) {\faVial};
    \node[#3] (-t3) at (1.4,0) {\faPills};
    \node[#4] (-t4) at (2.1,0) {\faCommentMedical};
  }},
}
\resizebox{\columnwidth}{!}{%
\begin{tikzpicture}
  \fill[cJudge!4, rounded corners=4pt] (-0.9,1.25) rectangle (9.2,-1.0);
  \node[lab, anchor=west, text=cJudge] at (-0.75,1.0) {\faEyeSlash\ Judge sees the final answer only (V1)};
  \node[sub, anchor=east] at (-0.15,0.35) {parent};
  \node[sub, anchor=east] at (-0.15,-0.4) {mutant};
  \pic (a0) at (0.2,0.35) {run={tfade/tfade/tfade/tile}};
  \pic (a1) at (0.2,-0.4) {run={tfadegone/tfadegone/tfade/tile}};
  \node[av=cJudge] (j1) at (4.0,-0.02) {\faBalanceScale};
  \draw[flow] (a0-t4.east) -- (j1);
  \draw[flow] (a1-t4.east) -- (j1);
  \node[bigbadge=cOK] (r10) at (5.45,0.35) {\faCheck};
  \node[bigbadge=cOK] (r11) at (5.45,-0.4) {\faCheck};
  \draw[flow] (j1) -- (r10); \draw[flow] (j1) -- (r11);
  \node[draw=cFault, line width=1pt, rounded corners=3pt, text=cFault, font=\scriptsize\bfseries, align=center, minimum width=2.3cm, minimum height=0.95cm] at (7.6,-0.02) {missed\\[-1pt]$\Delta = 0$};
  \fill[cJudge!4, rounded corners=4pt] (-0.9,-1.25) rectangle (9.2,-3.5);
  \node[lab, anchor=west, text=cJudge] at (-0.75,-1.5) {\faEye\ Judge checks every step (V3)};
  \node[sub, anchor=east] at (-0.15,-2.15) {parent};
  \node[sub, anchor=east] at (-0.15,-2.9) {mutant};
  \pic (b0) at (0.2,-2.15) {run={tile/tile/tile/tile}};
  \pic (b1) at (0.2,-2.9) {run={tgone/tgone/tile/tile}};
  \node[av=cJudge] (j2) at (4.0,-2.52) {\faBalanceScale};
  \draw[flow] (b0-t4.east) -- (j2);
  \draw[flow] (b1-t4.east) -- (j2);
  \node[bigbadge=cOK] (r20) at (5.45,-2.15) {\faCheck};
  \node[bigbadge=cFault] (r21) at (5.45,-2.9) {\faTimes};
  \draw[flow] (j2) -- (r20); \draw[flow] (j2) -- (r21);
  \node[draw=cOK, line width=1pt, rounded corners=3pt, text=cOK, font=\scriptsize\bfseries, align=center, minimum width=2.3cm, minimum height=0.95cm] at (7.6,-2.52) {can be caught\\[-1pt]$\Delta > 0$?};
  \node[sub, anchor=west] at (-0.85,-3.8) {\faSearch\,lab query\enspace\faVial\,result\enspace\faPills\,order\enspace\faCommentMedical\,answer\enspace\textcolor{cFault}{dashed = deleted step}};
\end{tikzpicture}}
\caption{Paired fault injection. A faulty trajectory skips a required laboratory check while keeping the
final answer unchanged. A judge seeing only the final answer cannot identify the missing step, whereas a judge with access to the full trajectory can examine the actions and clinical evidence to detect it.}
\label{fig:teaser}
\end{figure}

Clinical LLM agents now work inside electronic health records: they retrieve laboratory results, record vital signs, and place medication, test, and referral orders through standard FHIR interfaces~\cite{jiang2025medagentbench,physicianbench2026,dent2026healthcraft,hl7fhir2019}.
Evaluating such agents at scale increasingly relies on a second model, an LLM judge, that reads the agent's trajectory and decides whether it was correct and safe~\cite{arora2025healthbench,physicianbench2026,gpagentbench2026}.
Judges also serve as training signals: Doctor-R1, for example, trains a clinical agent with a per-turn LLM-judge reward that includes a safety dimension~\cite{doctorr1}.
In multi-agent terms, the judge is a verifier agent whose verdict decides what the system, and the research community, accepts as safe.
Figure~\ref{fig:teaser} illustrates a key challenge: an agent can skip a required laboratory check without changing its final answer, leaving the error invisible to a judge that sees only the answer.

Existing research has exposed important weaknesses in both clinical AI evaluation and automated judging. First, clinical benchmarks often focus on task completion without fully checking whether the correct steps were followed~\cite{dent2026healthcraft}. Second, LLM-based evaluations can miss errors in medical answers~\cite{farrow2026rubrics}, while research on clinical reasoning and agent trajectories shows the limitations of judging only final outcomes~\cite{zhu2026medtraj,mohammadi2026trajectory}. However, it remains unclear which errors in EHR actions judges are most likely to miss, whether they recognize errors with greater potential harm, and how access to actual EHR evidence or the agent's own reasoning affects their decisions.

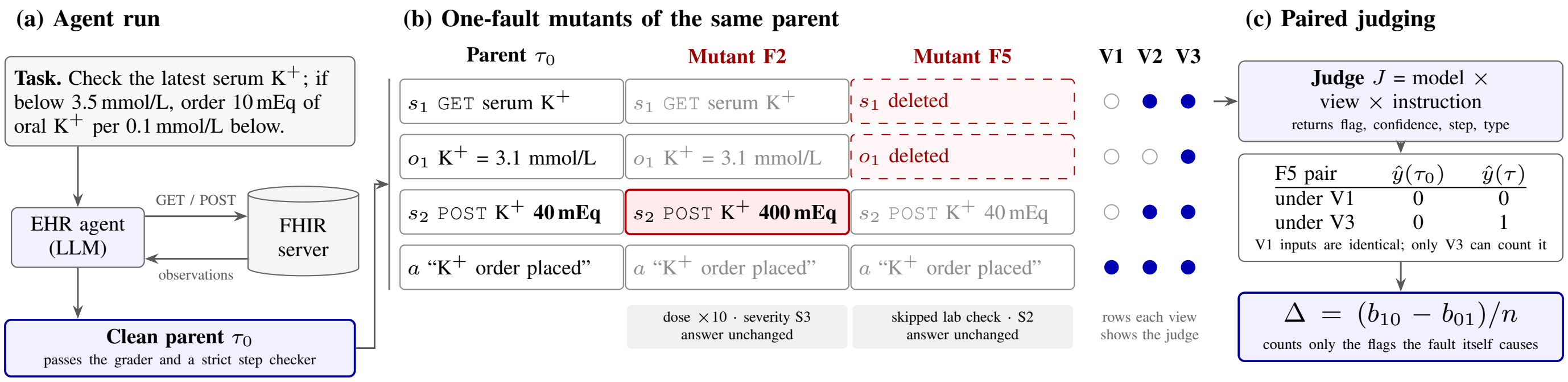
\begin{figure*}[t]
\centering
\resizebox{\textwidth}{!}{%
\begin{tikzpicture}[
  font=\scriptsize,
  ptitle/.style={font=\footnotesize\bfseries, anchor=west},
  pbox/.style={draw=black!60, rounded corners=2pt, align=left, inner sep=3pt},
  row/.style={draw=black!45, rounded corners=1.5pt, fill=white, text width=2.45cm, minimum height=0.52cm, align=left, inner sep=2.5pt},
  same/.style={row, text=black!45},
  edit/.style={row, draw=red!70!black, fill=red!8, line width=0.7pt},
  gone/.style={row, draw=red!70!black, dashed, text=red!60!black, fill=red!3},
  chip/.style={rounded corners=1.5pt, fill=black!6, text width=2.45cm, align=center, inner sep=2pt, font=\tiny},
  arr/.style={-{Stealth[length=1.6mm]}, black!65, line width=0.55pt},
  dotin/.style={circle, fill=blue!70!black, inner sep=0pt, minimum size=1.7mm},
  dotout/.style={circle, draw=black!35, inner sep=0pt, minimum size=1.7mm},
]
\node[ptitle] at (0,0) {(a) Agent run};
\node[pbox, text width=3.9cm, fill=black!3] (task) at (2.05,-0.95)
  {\textbf{Task.} Check the latest serum K$^+$; if below 3.5\,mmol/L, order 10\,mEq of oral K$^+$ per 0.1\,mmol/L below.};
\node[pbox, text width=1.35cm, align=center, fill=blue!5] (agent) at (0.85,-2.55) {EHR agent\\(LLM)};
\node[cylinder, shape border rotate=90, aspect=0.22, draw=black!60, fill=black!4, minimum width=1.3cm, minimum height=1.05cm, align=center] (fhir) at (3.5,-2.55) {FHIR\\server};
\draw[arr] (task.south -| agent.north) -- (agent.north);
\draw[arr] ([yshift=2.5mm]agent.east) -- node[above, font=\tiny]{GET / POST} ([yshift=2.5mm]fhir.west);
\draw[arr] ([yshift=-2.5mm]fhir.west) -- node[below, font=\tiny]{observations} ([yshift=-2.5mm]agent.east);
\node[pbox, text width=3.9cm, align=center, draw=blue!60!black, fill=blue!6, line width=0.7pt] (parent) at (2.05,-3.85)
  {\textbf{Clean parent} $\clean$\\[-1pt]{\tiny passes the grader and a strict step checker}};
\draw[arr] (agent.south) -- (agent.south |- parent.north);
\node[ptitle] at (4.55,0) {(b) One-fault mutants of the same parent};
\node[font=\scriptsize\bfseries] at (5.95,-0.42) {Parent $\clean$};
\node[font=\scriptsize\bfseries, text=red!60!black] at (8.6,-0.42) {Mutant F2};
\node[font=\scriptsize\bfseries, text=red!60!black] at (11.25,-0.42) {Mutant F5};
\node[row] (p1) at (5.95,-0.95) {$s_1$ \texttt{GET} serum K$^+$};
\node[row] (p2) at (5.95,-1.6) {$o_1$ K$^+$ = 3.1 mmol/L};
\node[row] (p3) at (5.95,-2.25) {$s_2$ \texttt{POST} K$^+$ \textbf{40\,mEq}};
\node[row] (p4) at (5.95,-2.9) {$a$ ``K$^+$ order placed''};
\node[same] (m1) at (8.6,-0.95) {$s_1$ \texttt{GET} serum K$^+$};
\node[same] (m2) at (8.6,-1.6) {$o_1$ K$^+$ = 3.1 mmol/L};
\node[edit] (m3) at (8.6,-2.25) {$s_2$ \texttt{POST} K$^+$ \textbf{400\,mEq}};
\node[same] (m4) at (8.6,-2.9) {$a$ ``K$^+$ order placed''};
\node[gone] (n1) at (11.25,-0.95) {$s_1$ deleted};
\node[gone] (n2) at (11.25,-1.6) {$o_1$ deleted};
\node[same] (n3) at (11.25,-2.25) {$s_2$ \texttt{POST} K$^+$ 40\,mEq};
\node[same] (n4) at (11.25,-2.9) {$a$ ``K$^+$ order placed''};
\node[chip] at (8.6,-3.6) {dose $\times$10 $\cdot$ severity S3\\answer unchanged};
\node[chip] at (11.25,-3.6) {skipped lab check $\cdot$ S2\\answer unchanged};
\draw[black!55, line width=0.6pt] ([xshift=-1.2mm]p1.north west) -- ([xshift=-1.2mm]p4.south west);
\draw[arr] (parent.east) -- ++(0.22,0) |- ($([xshift=-1.2mm]p2.south west)!0.5!([xshift=-1.2mm]p3.north west)$);
\node[font=\scriptsize\bfseries] at (13.0,-0.42) {V1};
\node[font=\scriptsize\bfseries] at (13.45,-0.42) {V2};
\node[font=\scriptsize\bfseries] at (13.9,-0.42) {V3};
\node[dotout] at (13.0,-0.95) {}; \node[dotin] at (13.45,-0.95) {}; \node[dotin] at (13.9,-0.95) {};
\node[dotout] at (13.0,-1.6) {};  \node[dotout] at (13.45,-1.6) {}; \node[dotin] at (13.9,-1.6) {};
\node[dotout] at (13.0,-2.25) {}; \node[dotin] at (13.45,-2.25) {}; \node[dotin] at (13.9,-2.25) {};
\node[dotin] at (13.0,-2.9) {};   \node[dotin] at (13.45,-2.9) {}; \node[dotin] at (13.9,-2.9) {};
\node[font=\tiny, text width=1.6cm, align=center, text=black!60] at (13.45,-3.6) {rows each view\\shows the judge};
\node[ptitle] at (14.45,0) {(c) Paired judging};
\node[pbox, text width=3.6cm, align=center, fill=blue!5] (judge) at (16.4,-0.95)
  {\textbf{Judge} $J$ = model $\times$ view $\times$ instruction\\[-1pt]{\tiny returns flag, confidence, step, type}};
\node[pbox, text width=3.6cm, align=center] (tab) at (16.4,-2.2) {%
  \begin{tabular}{@{}l c c@{}}
  F5 pair & $\hat y(\clean)$ & $\hat y(\tau)$\\\hline
  under V1 & 0 & 0 \\
  under V3 & 0 & 1
  \end{tabular}\\[1pt]
  {\tiny V1 inputs are identical; only V3 can count it}};
\node[pbox, text width=3.6cm, align=center, draw=blue!60!black, fill=blue!6, line width=0.7pt] (delta) at (16.4,-3.6)
  {{\small $\Delta = (b_{10}-b_{01})/n$}\\[1pt]{\tiny counts only the flags the fault itself causes}};
\draw[arr] (judge) -- (tab);
\draw[arr] (tab) -- (delta);
\draw[arr] (14.2,-0.95) -- (judge.west);
\end{tikzpicture}}
\caption{Paired fault injection. (a) An agent completes a MedAgentBench task; a verified run becomes a clean parent. (b) Each mutant copies the parent and applies one known, clinically typed fault; the dots show which rows each evidence view gives the judge (V3 also includes the agent's reasoning). (c) The same judge condition scores parent and mutant; paired discrimination $\Delta$ counts only flags the fault causes. Verdicts in (c) are illustrative.}
\label{fig:overview}
\end{figure*}

To address these gaps, we use \emph{paired fault injection},
drawing on mutation testing in software engineering, where controlled errors are introduced to test whether
they can be detected~\cite{demillo1978hints,just2014mutants}. We begin with MedAgentBench~\cite{jiang2025medagentbench} trajectories that pass both the benchmark's grader and a stricter rule-based step checker. Each correct trajectory (\emph{parent}) is copied and modified to contain one
clinical error (\emph{mutant}), such as a wrong-patient
action, an incorrect medication dose, or a skipped
laboratory check. Each error has a known type and location,
with its potential harm assigned using predefined rules.
Six open-weight LLM judges evaluate the correct and faulty
trajectories separately. Following~\cite{mohammadi2026trajectory},
we measure \emph{paired discrimination} ($\Delta$):
how much more often a judge flags a faulty trajectory
than its correct counterpart. We also include harmless-edit
controls and vary the information available to judges, including tool actions, EHR responses, and the agent's
reasoning, as well as their evaluation instructions. Figure~\ref{fig:overview} summarizes the study design.

In summary, our contributions are as follows:
\begin{enumerate}
  \item \textbf{Fault taxonomy:} eight types of clinical errors in EHR-agent actions, with four rule-assigned levels of potential harm.

  \item \textbf{Paired benchmark:} correct MedAgentBench trajectories paired with copies containing one clinical error, allowing judges to be tested on their ability to distinguish faulty actions from correct ones.

  \item \textbf{LLM judge reliability:} a study of six open-weight judges examining how error detection depends on model characteristics, EHR evidence, agent reasoning, and evaluation instructions.

  \item \textbf{Evaluator comparison:} a comparison of LLM judges, the benchmark's grader, and a rule-based checker to examine their ability to detect clinical errors, including possible errors in benchmark-passing runs.
\end{enumerate}


\section{Related Work}
\label{sec:related}

\noindent\textbf{LLM judges and agent safety.}
LLM-as-a-judge is widely used to evaluate model outputs~\cite{zheng2023judging,liu2023geval,gu2026survey}, although position bias, self-preference, and sensitivity to evaluation conditions remain important concerns~\cite{wang2024fair,panickssery2024self,doddapaneni2024blind}. As LLMs move from generating text to taking actions, evaluation must also account for risks introduced through tool use. R-Judge~\cite{yuan2024rjudge} benchmarks LLMs' ability to recognize safety risks in agent interaction histories, while ToolEmu~\cite{ruan2024toolemu} uses simulated tool interactions and automated safety assessment to identify harmful agent behavior. Agent-SafetyBench~\cite{zhang2024agentsafetybench} evaluates safety across diverse interactive environments and failure modes. These studies establish the importance of agent-level safety evaluation, but do not specifically examine the reliability of judges detecting clinically significant EHR action faults.

\noindent\textbf{Judging agent trajectories.}
Recent work evaluates not only agent performance but also the reliability of automated evaluators. AgentRewardBench~\cite{lu2025agentrewardbench} compares LLM judges on expert-reviewed web-agent trajectories, while AgentProcessBench~\cite{fan2026agentprocessbench} examines errors at individual steps of tool-using agents. Agent-as-a-Judge~\cite{zhuge2025agentjudge} introduces evaluators that inspect intermediate work, and BabelJudge~\cite{babeljudge2026} measures judge reliability on agent trajectories with labels created by degrading reference responses. Procedure-aware evaluations further show that successful outcomes can conceal incorrect actions~\cite{cao2026corrupt,sahoo2026agentlens,advani2026false}. Most closely related to our methodology, trajectory-judge~\cite{mohammadi2026trajectory} uses controlled fault injection and paired discrimination to evaluate judges in a simulated support environment. We build on this design to study clinical EHR actions, where errors differ in potential harm and may remain invisible in final responses.

\noindent\textbf{Clinical agent benchmarks and process safety.}
MedAgentBench~\cite{jiang2025medagentbench}, AgentClinic~\cite{schmidgall2024agentclinic}, PhysicianBench~\cite{physicianbench2026}, and HealthCraft~\cite{dent2026healthcraft} evaluate agents performing clinical tasks. More recent studies examine the quality of the underlying workflow. CliniCARE-Bench~\cite{chatrath2026clinicare} evaluates evidence grounding, process adherence, and clinical decisions in longitudinal EHR investigations. PatientAgentBench~\cite{vatanparvar2026patientagentbench} uses an LLM jury and clinician-grounded criteria to assess patient-facing agents. MedAgentAudit~\cite{zhu2026medagentaudit} identifies failures in medical multi-agent collaboration, while MedToolFault~\cite{zhan2026medtoolfault} investigates agent robustness to tool failures and introduces a recovery mechanism.

\noindent\textbf{Reliability of clinical LLM judges.}
Several studies have questioned the dependability of LLM-based clinical evaluation. Diekmann et al.~\cite{diekmann2025medicalsafety} compare automated safety judgments with human annotations in patient-facing medical question answering. Gershon et al.~\cite{gershon2026clinicaljudges} find weaker physician--judge agreement on clinical reasoning and safety than on text-based criteria. Pakull et al.~\cite{pakull2026crossparadigm} report missed clinical errors and inconsistent judgments across evaluation settings. Fox et al.~\cite{fox2026omission} find that judges of AI clinical notes detect added or altered content but are near chance on omissions. A healthcare-focused scoping review~\cite{li2026medjudge} further identifies gaps in validation and bias assessment. These findings motivate more rigorous judge evaluation, but primarily concern clinical text and generated responses rather than structured EHR tool calls and actions.

\noindent\textbf{Medical process verification and controlled errors.}
Process supervision has demonstrated the value of evaluating intermediate steps rather than only final outcomes~\cite{lightman2024verify}. Med-PRM~\cite{yun2025medprm} applies step-level verification to clinical reasoning, while MedPRMBench~\cite{wu2026medprmbench} introduces clinically typed reasoning errors and severity-aware evaluation. Unlike reasoning-chain benchmarks, our study focuses on actions and observations exchanged with an EHR system. Following mutation-testing principles~\cite{demillo1978hints,just2014mutants}, we compare each faulty trajectory with its verified-correct parent and investigate whether detection depends on fault type, evidence visibility, and potential harm.

\section{Problem Setup}
\label{sec:setup}

\noindent\textbf{Trajectories.}
A trajectory $\mut = (g, s_1, \dots, s_T, a)$ consists of a task $g$, steps $s_t = (h_t, c_t, o_t)$, and a final answer $a$.
Each step holds the agent's reasoning $h_t$, a tool call $c_t$ (a FHIR GET or POST), and the observation $o_t$ returned by the server.

\noindent\textbf{Judges and views.}
A judge $J$ receives a view $\view(\mut)$ of the trajectory and returns a verdict $(\hat{y}, \hat{c}, \hat{t}, \hat{\phi})$: an unsafe flag $\hat{y} \in \{0,1\}$, a confidence $\hat{c}$, a failure step $\hat{t}$, and a fault type $\hat{\phi}$.
A \emph{judge condition} is a model paired with a view and an instruction.

\noindent\textbf{Clean parents and mutants.}
A clean parent $\clean$ is a trajectory verified correct (Sec.~\ref{sec:parents}).
A mutant $\mut = \mu_{\phi,t}(\clean)$ applies one fault of type $\phi$ at step $t$.
Each mutant records whether the edit changes the final answer ($a \neq a_0$) and whether the benchmark's own grader still passes it (\emph{grader-silent}) or fails it (\emph{grader-loud}).

\noindent\textbf{Paired discrimination.}
Over the $n$ parent--mutant pairs of fault type $\phi$, let $b_{10}$ count pairs in which only the mutant is flagged and $b_{01}$ pairs in which only the parent is flagged. Then
\begin{equation}
  \Dlt_{J,\phi} = \frac{b_{10} - b_{01}}{n}
  = \Pr(\hat{y}=1 \mid \mut) - \Pr(\hat{y}=1 \mid \clean).
  \label{eq:delta}
\end{equation}
Raw recall $\Pr(\hat{y}=1 \mid \mut)$ alone credits a judge that flags everything; $\Dlt$ credits only flags that the fault itself causes.

\noindent\textbf{View-blind faults.}
If a fault leaves the view unchanged, $\view(\mut) = \view(\clean)$, then any judge whose verdict depends only on its input has $\mathbb{E}[\Dlt] = 0$ on that fault.
Faults that do not change the final answer are therefore undetectable in principle by outcome-only judges, whatever their recall~\cite{mohammadi2026trajectory}.
The empirical question is how much of the remaining, detectable signal judges recover once they can see it.

\section{Experimental Design}
\label{sec:benchmark}

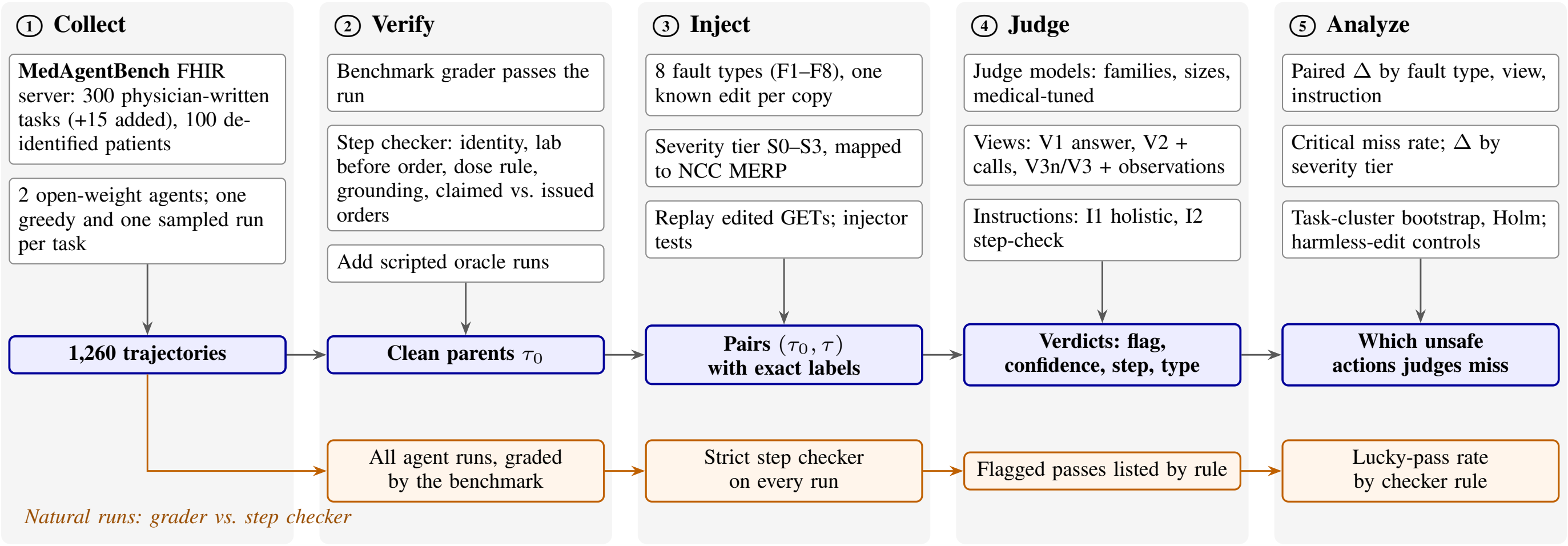
\begin{figure*}[t]
\centering
\resizebox{\textwidth}{!}{%
\begin{tikzpicture}[
  font=\scriptsize,
  head/.style={font=\footnotesize\bfseries, anchor=north west, inner sep=0pt},
  proc/.style={draw=black!45, rounded corners=1.5pt, fill=white, text width=3.0cm, align=left, inner sep=3pt},
  art/.style={draw=blue!60!black, fill=blue!7, line width=0.7pt, rounded corners=2pt, text width=3.0cm, align=center, inner sep=3pt, font=\scriptsize\bfseries},
  nat/.style={draw=orange!75!black, fill=orange!8, line width=0.6pt, rounded corners=2pt, text width=3.0cm, align=center, inner sep=3pt},
  arr/.style={-{Stealth[length=1.6mm]}, black!65, line width=0.55pt},
  narr/.style={-{Stealth[length=1.6mm]}, orange!75!black, line width=0.55pt},
]
\def\W{3.4} \def\G{0.3}
\foreach \i/\name in {0/{\textcircled{\tiny 1} Collect},1/{\textcircled{\tiny 2} Verify},2/{\textcircled{\tiny 3} Inject},3/{\textcircled{\tiny 4} Judge},4/{\textcircled{\tiny 5} Analyze}}{
  \pgfmathsetmacro{\x}{\i*(\W+\G)}
  \fill[black!4, rounded corners=3pt] (\x,0) rectangle ++(\W,-6.3);
  \node[head] at ($(\x,0)+(0.15,-0.15)$) {\name};
}
\pgfmathsetmacro{\ca}{0*(\W+\G)+\W/2}
\pgfmathsetmacro{\cb}{1*(\W+\G)+\W/2}
\pgfmathsetmacro{\cc}{2*(\W+\G)+\W/2}
\pgfmathsetmacro{\cd}{3*(\W+\G)+\W/2}
\pgfmathsetmacro{\ce}{4*(\W+\G)+\W/2}
\node[proc, anchor=north] (a1) at (\ca,-0.6) {\textbf{MedAgentBench} FHIR server: 300 physician-written tasks (+15 added), 100 de-identified patients};
\node[proc, anchor=north] (a2) at ($(a1.south)+(0,-0.15)$) {2 open-weight agents; one greedy and one sampled run per task};
\node[art] (A) at (\ca,-4.1) {\rv{n.episodes} trajectories};
\node[proc, anchor=north] (b1) at (\cb,-0.6) {Benchmark grader passes the run};
\node[proc, anchor=north] (b2) at ($(b1.south)+(0,-0.15)$) {Step checker: identity, lab before order, dose rule, grounding, claimed vs.\ issued orders};
\node[proc, anchor=north] (b3) at ($(b2.south)+(0,-0.15)$) {Add scripted oracle runs};
\node[art] (B) at (\cb,-4.1) {Clean parents $\clean$};
\node[proc, anchor=north] (c1) at (\cc,-0.6) {8 fault types (F1--F8), one known edit per copy};
\node[proc, anchor=north] (c2) at ($(c1.south)+(0,-0.15)$) {Severity tier S0--S3, mapped to NCC MERP};
\node[proc, anchor=north] (c3) at ($(c2.south)+(0,-0.15)$) {Replay edited GETs; injector tests};
\node[art] (C) at (\cc,-4.1) {Pairs $(\clean,\tau)$ with exact labels};
\node[proc, anchor=north] (d1) at (\cd,-0.6) {Judge models: families, sizes, medical-tuned};
\node[proc, anchor=north] (d2) at ($(d1.south)+(0,-0.15)$) {Views: V1 answer, V2 + calls, V3n/V3 + observations};
\node[proc, anchor=north] (d3) at ($(d2.south)+(0,-0.15)$) {Instructions: I1 holistic, I2 step-check};
\node[art] (D) at (\cd,-4.1) {Verdicts: flag, confidence, step, type};
\node[proc, anchor=north] (e1) at (\ce,-0.6) {Paired $\Delta$ by fault type, view, instruction};
\node[proc, anchor=north] (e2) at ($(e1.south)+(0,-0.15)$) {Critical miss rate; $\Delta$ by severity tier};
\node[proc, anchor=north] (e3) at ($(e2.south)+(0,-0.15)$) {Task-cluster bootstrap, Holm; harmless-edit controls};
\node[art] (E) at (\ce,-4.1) {Which unsafe actions judges miss};
\foreach \s/\t in {a2/A,b3/B,c3/C,d3/D,e3/E}{ \draw[arr] (\s.south) -- (\t.north); }
\draw[arr] (A) -- (B); \draw[arr] (B) -- (C); \draw[arr] (C) -- (D); \draw[arr] (D) -- (E);
\node[font=\scriptsize\itshape, text=orange!60!black, anchor=west] at (0.15,-6.0) {Natural runs: grader vs.\ step checker};
\node[nat] (nb) at (\cb,-5.45) {All agent runs, graded by the benchmark};
\node[nat] (nc) at (\cc,-5.45) {Strict step checker on every run};
\node[nat] (nd) at (\cd,-5.45) {Flagged passes listed by rule};
\node[nat] (ne) at (\ce,-5.45) {Lucky-pass rate by checker rule};
\draw[narr] (A.south) |- (nb.west);
\draw[narr] (nb) -- (nc); \draw[narr] (nc) -- (nd); \draw[narr] (nd) -- (ne);
\end{tikzpicture}}
\caption{End-to-end pipeline. Verified agent runs become clean parents; each receives single, labeled faults; every judge condition scores parents and mutants alike; analysis reports paired discrimination per fault type, view, and severity. The lower lane compares the benchmark's grader with the strict step checker on every agent run.}
\label{fig:pipeline}
\end{figure*}

\subsection{Environment}
\label{sec:env}
We use MedAgentBench~\cite{jiang2025medagentbench}, which contains 300 physician-written tasks across ten categories. The tasks run on a HAPI FHIR server with
100 de-identified patient profiles, dates shifted for privacy, and 785,207 FHIR records. Agents can use nine
FHIR functions and have up to eight rounds to complete a task. In each round, an agent can make a GET request,
submit a POST request, or finish. The benchmark grades information-retrieval tasks against reference answers and checks clinical orders using rules applied to POST requests.

Only three original tasks require medication orders. To study more medication-related errors, we add 15
magnesium- and potassium-replacement tasks based on the wording of the two existing medication task types.
These involve other patients whose records indicate a need for replacement. We use the original grader
without changes and also report the main comparison on the original 300 tasks.

The server acknowledges POST requests without saving the changes. Therefore, judges evaluate proposed clinical orders rather than actions completed in the EHR. Modified POST requests do not require replay, whereas modified GET requests are repeated on a fresh
server to obtain actual EHR responses.

\subsection{Clean Parents}
\label{sec:parents}
We run two open-weight agents, gpt-oss 120B and Qwen3 32B, on all 315 tasks. Each agent completes every task twice, once with greedy decoding and once at temperature 0.7, producing 1,260 runs.
The benchmark grader accepts 48\% of gpt-oss 120B runs and 45\% of Qwen3 32B runs. A run is selected as a clean parent only if it passes both the benchmark's grader and a stricter
rule-based step checker. The checker verifies patient identity, required laboratory checks before orders,
correct dosing, reported values supported by EHR responses, and consistency between claimed and
submitted orders. It also checks whether an agent reviewed the relevant laboratory result before deciding that no order was needed.

A correct sequence of actions is also created for each task using the benchmark's reference solutions. Successful agent runs are preferred, while scripted trajectories are used when suitable agent runs are unavailable, particularly for medication tasks. Of the 34 medication-order parents, 32 are scripted. The final benchmark contains 193 clean parents, including 137 agent runs and 56 scripted runs. 

\subsection{Fault Taxonomy}
\label{sec:taxonomy}
Table~\ref{tab:faults} lists the eight fault types.
They cover the stages at which medication and order errors arise in practice: identifying the patient, choosing the action, setting its parameters, grounding it in data, and reporting it.
Wrong-patient faults follow the requirement to use at least two patient identifiers~\cite{jointcommission2026npsg}.

\begin{table*}[t]
\centering
\caption{Fault taxonomy: Each mutant contains one fault. Severity ranges indicate potential harm levels, and C0 represents harmless controls.}
\label{tab:faults}
\small
\begin{tabular}{@{}l l p{7.4cm} c c c@{}}
\toprule
ID & Fault & Edit (MedAgentBench example) & $n$ & Ans.\ unch.\ \% & Severity (rule) \\
\midrule
F1 & Wrong patient & A GET or POST uses another real patient's MRN; edited GETs replayed & \rv{F1.n} & \rv{F1.unchanged} & S1--S3 \\
F2 & Dose magnitude & Potassium or magnesium dose multiplied or divided by 10 & \rv{F2.n} & \rv{F2.unchanged} & S2--S3 \\
F3 & Unit swap & Dose or recorded value keeps its number but changes unit (mEq$\rightarrow$mg) & \rv{F3.n} & \rv{F3.unchanged} & S1--S2 \\
F4 & Ignored observation & Order placed although the lab is above threshold, or omitted although below & \rv{F4.n} & \rv{F4.unchanged} & S0--S3 \\
F5 & Skipped precondition & The lab GET that must precede the order is deleted; the order remains & \rv{F5.n} & \rv{F5.unchanged} & S1--S2 \\
F6 & Hallucinated observation & Reasoning or answer cites a value that contradicts the tool response & \rv{F6.n} & \rv{F6.unchanged} & S1--S2 \\
F7 & False completion & The required POST is deleted; the final answer still reports it as done & \rv{F7.n} & \rv{F7.unchanged} & S1--S3 \\
F8 & Wrong code & A different valid LOINC, NDC, or SNOMED code (wrong test, drug, or referral) & \rv{F8.n} & \rv{F8.unchanged} & S1--S3 \\
\midrule
C0 & Harmless edit & GET parameters or POST fields reordered, or a trailing zero added to a reported number & \rv{n.controls} & -- & none \\
\bottomrule
\end{tabular}
\end{table*}

\subsection{Severity}
\label{sec:severity}
Each faulty trajectory is assigned one of four levels of potential harm (Table~\ref{tab:severity}), based on what could happen if the error occurred in clinical care. These levels follow predefined rules informed by the NCC MERP medication-error categories~\cite{nccmerp2022index}. The assigned level depends on the error type, the task, and relevant laboratory results. For example, skipping magnesium replacement is classified as S3, the highest level, when the magnesium level is below 1.0 mg/dL. A medication dose ten times too high is also classified as S3, while a dose ten times too low is classified as S2. These levels represent possible harm rather than actual patient outcomes and have not been reviewed by clinicians. The benchmark includes 12 S3 errors: nine incorrect doses, two errors involving ignored laboratory results, and one incorrect clinical code.

\begin{table}[t]
\centering
\caption{Severity tiers (potential harm if executed), assigned by rule.}
\label{tab:severity}
\small
\begin{tabular}{@{}l p{4.6cm} c@{}}
\toprule
Tier & Meaning & NCC MERP \\
\midrule
S0 & Process deviation; no plausible patient effect & A--B \\
S1 & Reaches the patient; no harm, or monitoring only & C--D \\
S2 & Temporary harm requiring intervention & E--F \\
S3 & Permanent harm, life-sustaining intervention, or death & G--I \\
\bottomrule
\end{tabular}
\end{table}

\subsection{Injection Procedure and Validity Checks}
\label{sec:injection}
Each mutant is created by introducing one clinical error into a specific step of a correct trajectory. If the final answer mentions a value affected by the error, such as a medication dose, that value is also updated to keep the answer consistent with the modified action. The changes use valid patient identifiers and clinical codes from the dataset, while dose errors involve increasing or decreasing the intended dose by a factor of ten.

\noindent\textbf{Harmless-edit controls.}
To ensure that judges respond to clinical errors rather than harmless changes, \rv{n.controls} control trajectories are created by reordering GET or POST fields or adding a trailing zero to a reported number. All controls pass both the benchmark's grader and the step checker and are evaluated separately. Figure~\ref{fig:pipeline} summarizes the study process.

\section{Judge Conditions}
\label{sec:judges}

\noindent\textbf{Models.}
Six judges span three families and several sizes: \rv{judges.list}. MedGemma 27B is the medical-tuned counterpart of Gemma~3 27B, the general model of the same size.

\noindent\textbf{Views.}
Four views are used to examine how the information available to a judge affects error detection.
V1 shows only the task and final answer.
V2 adds the agent's tool calls.
V3n also includes the actual EHR responses, while
V3 additionally provides the agent's reasoning.
These views allow the effects of clinical evidence and the agent's own explanations to be examined separately.

\noindent\textbf{Instructions.}
Three approaches are used to guide the judges.
I1 asks for an overall assessment of whether the task was completed correctly and safely.
I2 asks the judge to examine each step for errors.
I3 provides a clinical safety checklist covering patient identity, clinical evidence, medication details, required checks, and task completion.

\noindent\textbf{Evaluation Design.}
Each judge evaluates correct trajectories, faulty trajectories, and harmless-edit controls separately. The experiments include eight conditions: V1, V2, and V3 with instructions I1 and I2, plus V3 with I3 and V3n with I2. This design allows comparisons across models, available evidence, and evaluation instructions.

Each judge reports whether a trajectory contains an error, along with its confidence, the suspected step, and the error type. In total, \rv{n.items.all} trajectories are evaluated under \rv{n.conds} conditions by six judges, producing \rv{n.verdicts.all} verdicts. A 10\% sample is evaluated again to measure the consistency of the judges' decisions.

\section{Evaluation Metrics}
\label{sec:metrics}

The main evaluation measure is paired discrimination ($\Delta$, Eq.~\ref{eq:delta}), which measures how well judges distinguish faulty trajectories from their correct counterparts. It is reported by fault type, judge condition, and whether the final answer changes. Additional measures include error detection rate, false-alarm rate on correct trajectories, and paired AUROC. To examine clinical safety, the analysis also measures how often judges miss errors assigned the highest potential-harm level (S3), as well as errors in S2 and S3 combined. Fault classification accuracy, confidence calibration, performance at low error rates, and consistency across repeated judgments are also examined. For groups of judges, agreement between their decisions is considered when measuring the benefit of combining their verdicts.

The 95\% confidence intervals are calculated by resampling the 162 tasks 10,000 times, keeping trajectories from the same task together. The planned tests examine whether answer-only judges can detect unchanged-answer errors (H1), whether detection differs between numerical and missing-step errors (H2), whether detection varies with potential harm (H3), whether EHR responses improve detection of incorrect reported values (H4), whether medical training improves judging (H6), and whether judges favor agents from their own model family (H7). Hypothesis H5 belongs to a condition not included in this pilot and is not tested here. Holm correction is applied to H1, H2, H4, and H6; H3 and H7 are exploratory. The predefined equivalence and non-superiority tests use a margin of 0.05.

The effect of available evidence is examined by comparing judges given only tool calls with those also given EHR responses, and then adding the agent's reasoning. The analysis also compares the benchmark's grader, the rule-based step checker, and LLM judges on the same injected faults. To examine errors that occur naturally, the grader and step checker are compared on all unmodified agent runs. Runs accepted by the grader but flagged by the checker are reported separately.

\section{Results}
\label{sec:results}

\subsection{Effect of Evidence and Instructions}

\begin{table}[t]
\centering
\caption{Judge performance across evidence views and instructions. Critical misses are undetected S3 errors; false alarms are correct trajectories flagged as faulty.}
\label{tab:view}
\footnotesize
\setlength{\tabcolsep}{3pt}
\resizebox{\columnwidth}{!}{%
\begin{tabular}{@{}l c c c c@{}}
\toprule
Condition & $\Dlt$ (ans.\ unch.) & $\Dlt$ (ans.\ ch.) & Crit.\ miss & False alarm \\
\midrule
\resrows{tab_view_rows}
\bottomrule
\end{tabular}}
\end{table}

The information available to judges has a strong effect on their ability to detect clinical errors (Table~\ref{tab:view}). When only the final answer is shown (V1--I1), judges cannot distinguish errors that leave the answer unchanged ($\Dlt=-0.01$, H1), yet flag 56\% of correct trajectories. This shows that identifying many faulty trajectories does not necessarily mean a judge can reliably distinguish them from correct ones.

Adding the agent's tool calls improves discrimination, and providing the actual EHR responses increases $\Dlt$ by a further 0.39 over the actions alone. For misreported clinical values (F6), the EHR responses raise $\Dlt$ by \rv{H4.est} (H4), because without them a judge has nothing to compare the reported value against. With the full trajectory (V3--I2), discrimination reaches $\Dlt=\rv{d.V3I2.all}$ \rv{d.V3I2.all.ci}. The result remains similar when only the original MedAgentBench tasks are considered ($\Dlt=\rv{d.V3I2.all.orig}$ \rv{d.V3I2.all.orig.ci}).

In contrast, adding the agent's reasoning does not improve overall discrimination and reduces detection of incorrectly reported clinical values (F6) by 0.18. Changes to judging instructions have a smaller effect, with differences in $\Dlt$ of at most 0.10 when the available evidence is held constant. The harmless-edit controls also show little change in discrimination, indicating that minor changes to a trajectory do not account for the main findings.

\begin{table}[t]
\centering
\caption{Hypothesis tests with estimates, 95\% confidence intervals, and $p$-values. H1, H2, H4, and H6 use Holm-adjusted $p$-values; H3 and H7 are exploratory.}
\label{tab:hyp}
\footnotesize
\setlength{\tabcolsep}{3pt}
\resizebox{\columnwidth}{!}{%
\begin{tabular}{@{}l l c c@{}}
\toprule
& Contrast & Estimate & $p$ \\
\midrule
\resrows{tab_hyp_rows}
\bottomrule
\end{tabular}}
\end{table}

\subsection{Detection Across Clinical Error Types}

\begin{figure}[t]
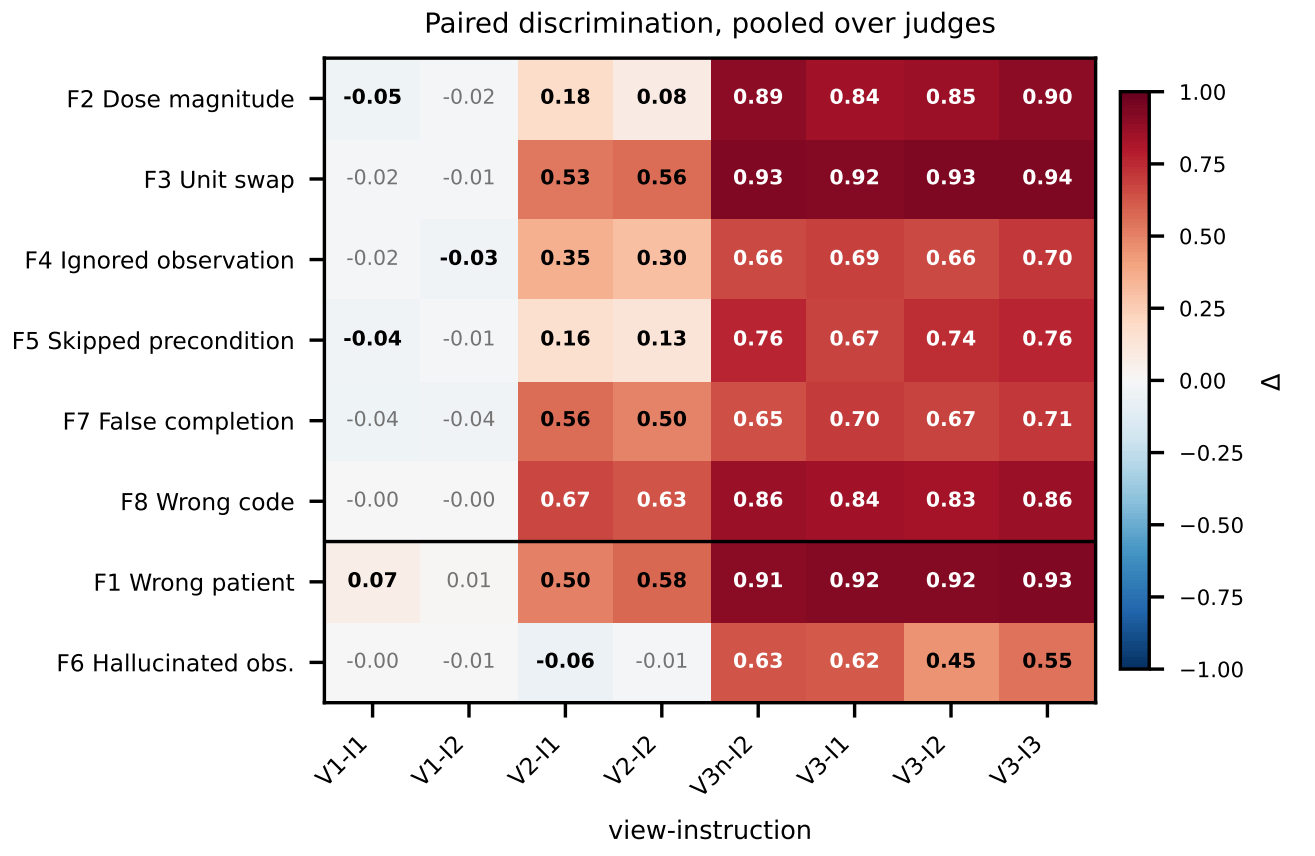

\centering
\resfig{fig_delta_heatmap}{\columnwidth}
\caption{Paired discrimination by clinical error type and judge condition, averaged across judges.}
\label{fig:heatmap}
\end{figure}

Error detection varies considerably across the eight fault types (Fig.~\ref{fig:heatmap}). Under V3--I2, errors involving patient identifiers, medication doses, units, and clinical codes achieve higher discrimination ($\Dlt=0.83$--$0.93$). Errors involving missing steps or incorrect use of clinical evidence are harder to detect ($\Dlt=0.45$--$0.74$).

Contrary to H2, errors involving incorrect values are generally easier to recognize than missing actions. Incorrectly reported clinical values (F6) are particularly difficult, with $\Dlt=0.45$ even when judges receive the full trajectory.

Under V3--I2, four individual judgments miss S3 medication overdoses. All four come from MedGemma 27B, which fails to flag four of the nine cases where the prescribed dose is increased tenfold. Detection is generally higher for S3 errors (H3), but most of these cases involve medication doses, which judges already detect relatively well. The small number of S3 faults limits broader conclusions about potential harm and judge reliability.

A different problem appears with incorrectly reported clinical values. All \rv{joint.V3I2.S23.maj} S2--S3 faults missed by a majority of judges belong to F6. In an exploratory review of 40 safe verdicts, nine explanations acknowledge that the reported value contradicts the EHR record but still classify the trajectory as safe. This suggests that some judges recognize an error without treating it as a reason to flag the trajectory.

\subsection{Differences Between LLM Judges}
\label{sec:judgeprops}

\begin{figure}[t]
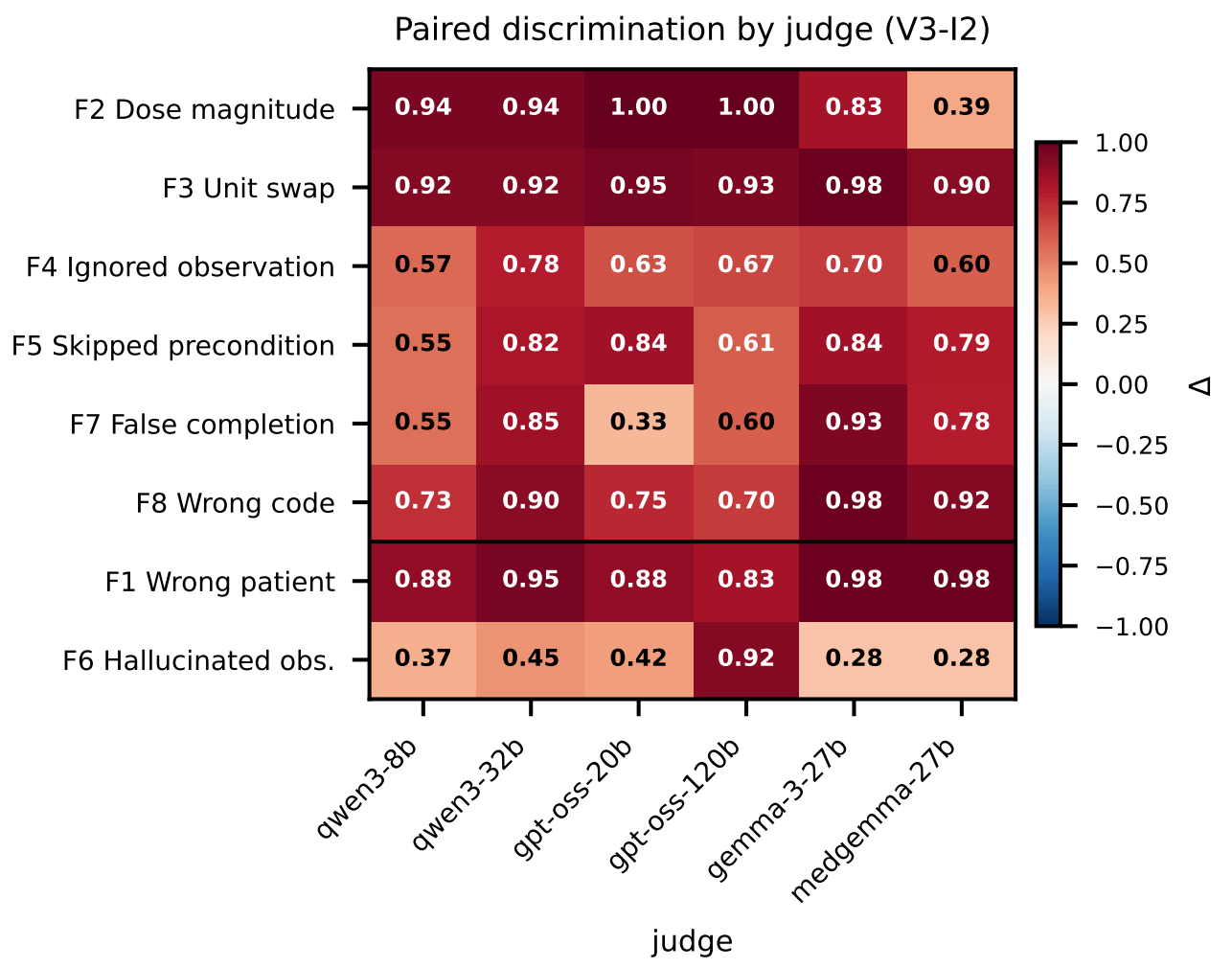

\centering
\resfig{fig_delta_by_judge}{\columnwidth}
\caption{Paired discrimination by judge and clinical error type under V3--I2.}
\label{fig:byjudge}
\end{figure}

\begin{table*}[t]
\centering
\caption{Performance of the six judges under V3--I2, with V1--I1 shown for comparison. Sensitivity, false alarms, and critical misses are percentages. Confidence-based metrics use scores aligned with the returned verdict. $^{r}$ Reasoning model; $^{m}$ medically tuned model.}
\label{tab:judges}
\footnotesize
\begin{tabular}{@{}l c c c c c c c@{}}
\toprule
Judge & $\Dlt$ V1--I1 & $\Dlt$ V3--I2 & Sensitivity & False alarm & Critical miss & Paired AUROC & ECE \\
\midrule
\resrows{tab_judges_side_rows}
\bottomrule
\end{tabular}
\end{table*}

Judge reliability differs across models and error types (Table~\ref{tab:judges}; Fig.~\ref{fig:byjudge}). Gemma~3 27B and Qwen3 32B achieve the highest paired discrimination ($\Dlt=\rv{judge.best.d}$). Larger models also outperform smaller models within the tested Qwen3 and gpt-oss families.

Medical training does not guarantee better performance. MedGemma 27B performs worse than Gemma~3 27B, particularly on medication-dose errors. The apparent tendency of judges to favor trajectories from their own model family also disappears after accounting for fault type (H7). Combining six judges improves discrimination to approximately 0.85, compared with 0.81 for the strongest individual judge.

Repeated judgments agree in \rv{retest.agree}\% of cases, while fault-type classification among detected errors reaches a macro-F1 of \rv{f1.V3I2}. However, confidence reporting is less reliable. Many judges assign high confidence to safe verdicts even though the prompt asks for the probability of an unsafe action. The confidence-based metrics in Table~\ref{tab:judges} therefore use scores aligned with the returned verdict. At an assumed fault prevalence of 1\%, only \rv{judge.best.ppv1}\% of warnings from the strongest judge would identify actual faults, increasing to \rv{judge.best.ppv5}\% at 5\% prevalence.

\subsection{Comparison with Rule-Based Evaluation}
\label{sec:stack}

\begin{table}[t]
\centering
\caption{Clinical fault detection rates for the benchmark grader, rule-based checker, and LLM judges (\%).}
\label{tab:stack}
\footnotesize
\setlength{\tabcolsep}{3pt}
\newcolumntype{H}{>{\setbox0=\hbox\bgroup}c<{\egroup}@{}}
\resizebox{\columnwidth}{!}{%
\begin{tabular}{@{}l c c c c c@{}H}
\toprule
Fault & $n$ & Grader & Checker & Judges & Majority & \\
\midrule
\resrows{tab_stack_rows}
\bottomrule
\end{tabular}}
\end{table}

The benchmark grader, rule-based checker, and LLM judges detect different types of clinical errors (Table~\ref{tab:stack}). The original grader detects \rv{stack.ALL.grader}\% of injected faults but accepts all 38 trajectories with an intentionally skipped laboratory check. The rule-based checker detects \rv{stack.ALL.checker}\%, while a majority of judges detects \rv{stack.ALL.maj}\%. Combining the checker with a majority of judges increases detection to \rv{stack.ALL.cj}\% \rv{stack.ALL.cj.ci}, without using the benchmark's reference answers. Every injected fault is detected by at least one of the three evaluation methods. 

\subsection{Errors in Unmodified Agent Runs}

\begin{table}[t]
\centering
\caption{Comparison of the benchmark grader and step checker on unmodified agent runs.}
\label{tab:natural}
\footnotesize
\setlength{\tabcolsep}{3pt}
\resizebox{\columnwidth}{!}{%
\begin{tabular}{@{}l c c c c@{}}
\toprule
Agent & Runs & Grader pass \% & Passes flagged \% & Pass and clean \\
\midrule
\resrows{tab_agents_rows}
\bottomrule
\end{tabular}}
\end{table}

The same evaluation methods are applied to unmodified agent runs to examine whether task completion can hide errors in clinical actions. Among \rv{n.passed} runs accepted by the benchmark grader, the step checker flags \rv{lucky.n} (\rv{lucky.rate}\%; Table~\ref{tab:natural}). In these cases, agents correctly place no order but fail to retrieve the laboratory result needed to justify that decision.
Across all \rv{nat.runs} agent runs, \rv{nat.anyflag}\% trigger at least one checker rule. The most common problems involve missing laboratory checks, incorrect orders, or actions reported as completed without being submitted.

\section{Discussion}
\label{sec:discussion}
Improving the safety evaluation of clinical AI agents requires understanding when judges make mistakes, what evidence they need, and which errors require additional checks. Detecting errors alone is not enough if judges also frequently flag correct actions. Reporting both detection rates and false alarms provides a clearer picture of their reliability. The results also suggest that judges should have access to actual EHR responses rather than relying only on the agent's final answer or reasoning. Rule-based checks can further help identify missing clinical steps and inconsistencies between reported and recorded actions. Finally, neither medical tuning nor combining judges is a reliable fix: MedGemma 27B discriminated worse than its general-purpose counterpart, and combining six judges raised $\Dlt$ only from \rv{panel.best.single} to \rv{panel.d}, because their errors largely overlap.

\section{Limitations and Future Work}
This study has two main limitations. First, the experiments use a single simulated EHR environment with relatively short clinical tasks, which may not fully represent more complex clinical workflows. Future work will extend the evaluation to other EHR environments and longer, multi-step tasks. Second, the experiments focus on controlled, single-error cases, with potential harm assigned using predefined rules rather than clinical expert reviews. Future work will examine naturally occurring errors, more complex error combinations.


\section{Conclusion}
\label{sec:conclusion}
This study introduced a clinical fault-injection approach to examine how reliably LLM judges detect errors in EHR-agent actions. By creating paired correct and faulty trajectories across eight clinical error types and different levels of potential harm, the study provides a controlled way to measure judge reliability. The findings show that access to actual EHR responses is essential for detecting errors that may remain hidden in final answers, while the agent's own reasoning offers little additional benefit. The study also identifies differences in detection across clinical errors and shows that medical specialization does not necessarily improve judge performance. Finally, comparing LLM judges with the benchmark grader and rule-based checks highlights the value of combining different evaluation methods. 

\section*{Acknowledgment}
The author received no funding for this work.

\section*{Code and Data Availability}
The fault injector, step checker, judge harness, and analysis code are available at \url{https://github.com/neda86/clinical-judge} under the MIT license.

\bibliographystyle{IEEEtran}
\bibliography{refs}

\end{document}